\documentclass[journal]{IEEEtran}
\usepackage{booktabs}
\usepackage{multirow}
\usepackage{makecell}
\usepackage{enumitem}
\usepackage{amsmath,amsfonts}
\usepackage{array}
\usepackage[caption=false,font=normalsize,labelfont=sf,textfont=sf]{subfig}
\usepackage{textcomp}
\usepackage{stfloats}
\usepackage{url}
\usepackage{verbatim}
\usepackage{graphicx}
\usepackage{cite}
\usepackage[table,xcdraw,x11names]{xcolor}
\usepackage{amssymb}
\usepackage{pifont}
\definecolor{commgreen}{HTML}{009933}
\usepackage[linesnumbered,ruled,vlined]{algorithm2e}

\SetCommentSty{licommfont}
\newcommand{\samplfull}{Pseudocode for RSD}
\definecolor{myorange}{RGB}{255,175,80}
\begin{document}

\title{LightLoc++: Sensor-Robust Representation Learning for Efficient Outdoor LiDAR Localization}

\author{Wen Li$^\ast$, Shangshu Yu$^\ast$, Dunqiang Liu, Shaoyang Chen, Qiming Xia, Sheng Ao,~\IEEEmembership{Member,~IEEE}, \\
Siqi Shen,~\IEEEmembership{Senior Member,~IEEE}, 
Chenglu Wen,~\IEEEmembership{Senior Member,~IEEE} and Cheng Wang,~\IEEEmembership{Senior Member,~IEEE}
\thanks{Manuscript received xxxx, 2025; revised xxxx, 2025, accepted xxxx, 2025.
This work was supported in part by the XXX. The Associate Editor for this paper was xxxx. (Corresponding author: Cheng Wang)

Wen Li, Dunqiang Liu, Shaoyang Chen, Qiming Xia, Sheng Ao, Siqi Shen, Chenglu Wen, and Cheng Wang are with the Fujian Key Laboratory of Urban Intelligent Sensing and Computing, and with Key Laboratory of Multimedia Trusted Perception and Efficient Computing, Ministry of Education of China, School of Informatics, Xiamen University, 422 Siming Road South, Xiamen, FJ 361005, P.R. China. Wen Li is also with the School of Engineering Mathematics and Technology, University of Bristol, Bristol BS8 1TH, United Kingdom. (e-mail: wen.li.sd95@gmail.com; dqliu@stu.xmu.edu.cn; sychan@stu.xmu.edu.cn; qimingxia@stu.xmu.edu.cn; aosh@xmu.edu.cn; siqishen@xmu.edu.cn; clwen@xmu.edu.cn; cwang@xmu.edu.cn).

Shangshu Yu is with the School of Computer Science and Engineering, Northeastern University, Shenyang 110819, China. (e-mail: yushangshu@cse.neu.edu.cn).

$^\ast$ Equal contribution.}}

\markboth{Journal of \LaTeX\ Class Files,~Vol.~14, No.~8, August~2021}%
{Shell \MakeLowercase{\textit{et al.}}: A Sample Article Using IEEEtran.cls for IEEE Journals}

\maketitle
\begin{abstract}
Scene coordinate regression (SCR) achieves strong performance in outdoor LiDAR localization, but usually requires days of scene-specific training, limiting its practicality for time-sensitive deployment. 
Recent decoupled methods improve efficiency by pretraining a scene-agnostic backbone on source datasets, freezing it for new scenes, and optimizing only lightweight scene-specific heads.
However, this paradigm heavily depends on the generalization ability of the pretrained backbone. Existing decoupled methods can match fully optimized SCR methods when target LiDAR configurations are similar to those used during backbone pretraining, but show a much larger performance gap on datasets collected with different LiDAR configurations. 
This suggests that the pretrained backbone is required to learn sensor-robust representations that capture stable scene geometry across LiDAR configurations.
Motivated by this observation, we propose LightLoc++, a sensor-robust and efficient outdoor LiDAR localization framework. 
To support sensor-robust representation learning, we introduce SULID, a synchronized urban multi-LiDAR dataset with representative 32-, 64-, and 128-beam rotating LiDARs, extensive cross-sensor overlap, and diverse urban scenes. Using SULID, we pretrain a sensor-robust backbone through cross-sensor consistency learning.
LightLoc++ preserves efficient new-scene learning by incorporating sample classification guidance and redundant sample downsampling, which reduce regression ambiguity and computational redundancy in large-scale outdoor scenes. 
Extensive experiments on multiple outdoor LiDAR localization benchmarks demonstrate that LightLoc++ achieves state-of-the-art localization performance while maintaining highly efficient new-scene training. 
Code and dataset will be made available at \url{https://github.com/liw95/LightLoc-PlusPlus}.
\end{abstract}

\section{Introduction}
\label{sec:intro}

LiDAR localization aims to estimate the 6-DoF pose of sensors, which is a fundamental component of many applications, \emph{e.g.,} autonomous driving~\cite{Yin_2024_IJCV} and robotics~\cite{Zhang_2024_AAAI}.

\begin{figure}[t]
  \centering
  \includegraphics[width=1\columnwidth]{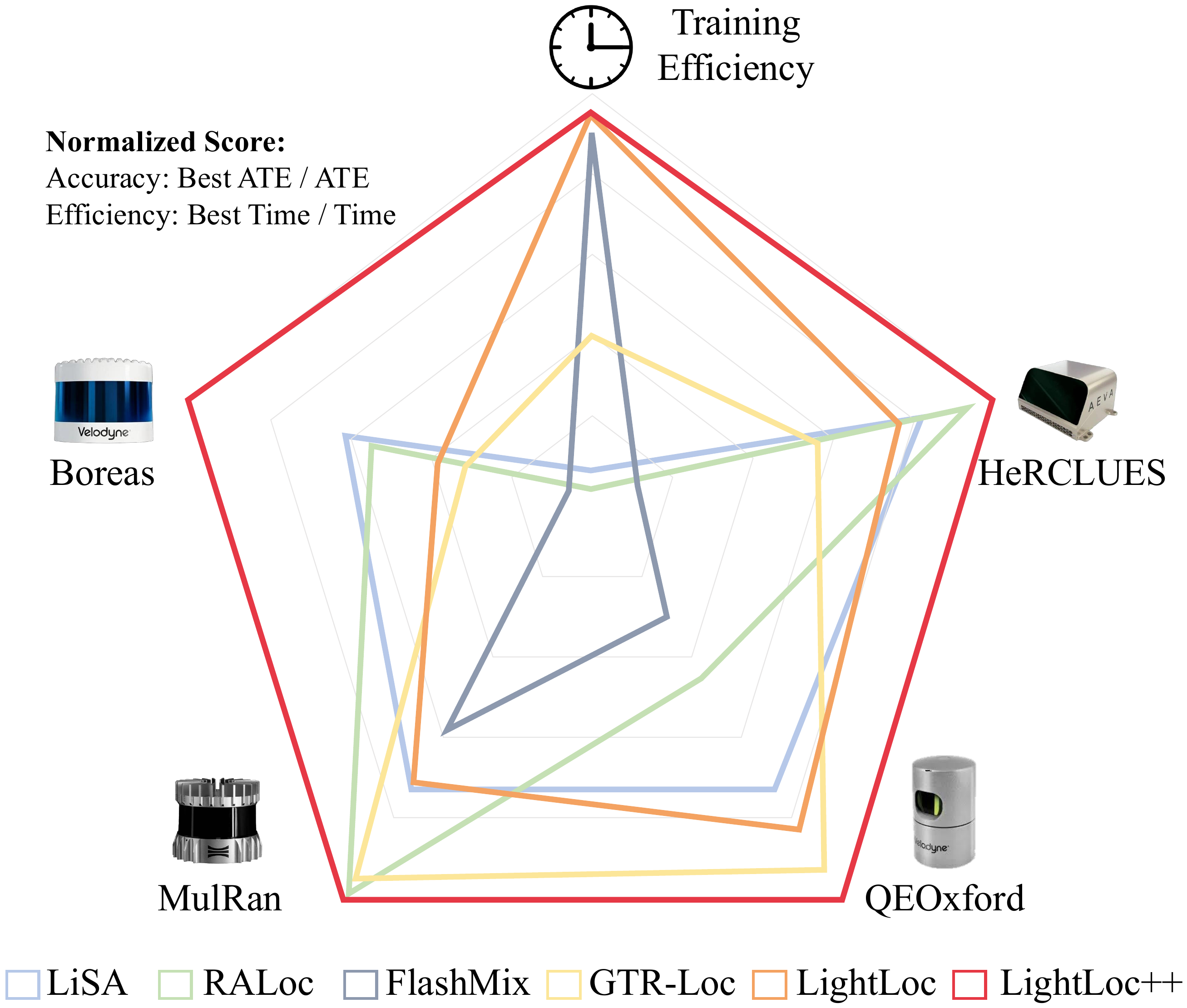}
   \caption{
   \textbf{LiDAR localization performance vs. training efficiency.} 
   The figure compares normalized localization accuracy and training efficiency across multiple LiDAR benchmarks collected with different sensors~\cite{burnett2023boreas,kim2020mulran,Dan_2020_ICRA,Li_2023_CVPR,kim2025hercules}. Larger values indicate better performance. LightLoc++ achieves consistently strong localization performance across diverse sensing modalities while maintaining high training efficiency.
   }
   \label{fig:1}
\end{figure}

Contemporary state-of-the-art LiDAR-based localization methods rely on explicit map usage, matching query points with a pre-built 3D map~\cite{Xia_2021_CVPR, Xia_2023_ICCV, Wang_2024_AAAI,yin2025general,luo2025bevplace++}.
However, these approaches often require costly map storage and incur significant communication overhead. 
Therefore, regression-based methods~\cite{Brahmbhatt_2018_CVPR, Chen_2024_CVPR, Li_2024_CVPR, Brachmann_2023_CVPR, Yang_2024_CVPR} have been proposed to address these limitations by memorizing the specific scene within a network. 
During inference, these methods eliminate the need for explicit map matching, thus reducing both storage and communication requirements.  
Based on their regression objectives, these methods can be categorized into Absolute Pose Regression (APR)~\cite{Brahmbhatt_2018_CVPR, Chen_2024_CVPR, Li_2024_CVPR} and Scene Coordinate Regression (SCR)~\cite{Brachmann_2023_CVPR, Li_2023_CVPR, Yang_2024_CVPR}.
Specifically, APR directly regresses poses, while SCR predicts point-to-world correspondences and estimates poses via RANSAC~\cite{Fischler_1981_Commun}. 
By explicitly leveraging geometric information, SCR achieves higher localization accuracy.

Despite their effectiveness, due to training needs to be repeated for each new scene, long training times make it impractical for applications. For example, LiSA~\cite{Yang_2024_CVPR}, a recent SCR-based localization method achieving 0.95m accuracy on the QEOxford dataset~\cite{Dan_2020_ICRA, Li_2023_CVPR}, requires approximately 53 hours of training as reported in~\cite{Li_2025_CVPR}. 
To reduce this cost, several methods~\cite{Yu_2022_PR,Goswami_2025_WACV,Li_2025_CVPR} decouple LiDAR localization into a scene-agnostic backbone and scene-specific prediction heads, where the backbone is pretrained on source datasets and frozen for new scenes, and only lightweight heads are optimized. For example, FlashMix~\cite{Goswami_2025_WACV} pretrains the backbone on MulRan~\cite{kim2020mulran} and Apollo-SouthBay~\cite{lu_2019_l3}, while LightLoc~\cite{Li_2025_CVPR} uses a backbone pretrained on nuScenes~\cite{Caesar_2020_CVPR}. This decoupled paradigm substantially improves the efficiency of new-scene learning.

However, the effectiveness of this decoupled paradigm critically depends on the generalization ability of the pretrained backbone.
As shown in Fig.~\ref{fig:1}, LightLoc achieves localization accuracy comparable to LiSA and RALoc~\cite{Yang_2025_ICCV} on QEOxford~\cite{Dan_2020_ICRA,Li_2023_CVPR}, where the LiDAR configuration is similar to that used for backbone pretraining. However, when evaluated on benchmarks collected with different LiDAR configurations, such as Boreas~\cite{burnett2023boreas}, MulRan~\cite{kim2020mulran}, and HeRCULES~\cite{kim2025hercules}, the performance gap becomes much larger. This indicates that existing pretrained backbones have limited generalization ability under sensor changes, which becomes a key bottleneck for efficient LiDAR localization.

An intuitive way to mitigate this bottleneck is to pretrain the backbone with data collected from multiple LiDAR sensors. This exposes the backbone to diverse LiDAR configurations and can improve its generalization ability. However, as LiDAR hardware continues to evolve, it is impractical to exhaustively collect data for every sensor type or retrain the backbone whenever a new sensor becomes available. 
We argue that generalization should not rely solely on increasing sensor coverage during training. Instead, the backbone should learn sensor-robust representations that capture stable scene geometry while reducing the influence of sensor-specific configurations.

Learning sensor-robust representations benefits from observing the same scene with multiple LiDAR sensors, as it allows the model to associate different sensor measurements with the same underlying physical structures.
Although recent datasets such as HeLiPR~\cite{jung_2024_helipr} and GEODE~\cite{chen2026heterogeneous} provide synchronized multi-LiDAR measurements, their effective cross-sensor overlap is often restricted by sensor placement, field-of-view differences, and occlusions. This limits the availability of rich cross-sensor correspondences over large scene regions. Moreover, these datasets provide limited coverage of diverse large-scale urban layouts.
These limitations motivate us to collect a new dataset specifically designed for sensor-robust backbone pretraining in efficient outdoor LiDAR localization.

We introduce SULID, a \textbf{S}ynchronized \textbf{U}rban mu\textbf{L}t\textbf{I}-Li\textbf{D}AR dataset for sensor-robust representation learning. 
Benefiting from the carefully designed sensor layout, SULID provides near-360$^\circ$ shared observations among three representative rotating LiDAR sensors, \emph{i.e.}, 32-, 64-, and 128-beam LiDARs, while preserving their different LiDAR configurations. This enables rich cross-sensor correspondences over most surrounding structures, making SULID well suited for learning sensor-robust representations.
In addition, SULID is collected across multiple urban regions and covers a wide range of scene types, including structured environments with regular layouts, open areas with weak geometric constraints, and dense urban areas with severe occlusions and complex scene structures.
By combining representative LiDAR configurations, near-panoramic shared observations, and diverse urban scene coverage, SULID goes beyond the specific localization task studied in this paper and can serve as a valuable resource for future research on sensor-robust perception and localization~\cite{yuan2024density,liu2026rcp,liu2025difflow3d}.

Based on SULID, we adopt a cross-sensor consistency learning strategy~\cite{xie2020pointcontrast,nunes2022segcontrast} to pretrain a sensor-robust backbone. Synchronized scans from the three LiDAR sensors are encoded by a shared backbone, where pooled scan-level features are aligned for global consistency and nearest-neighbor local features are aligned for local geometric consistency.

Beyond backbone pretraining, we further address efficient scene-specific training under the SCR framework. In related domains such as camera localization, prior works~\cite{Brachmann_2023_CVPR, Nguyen_2024_WACV, Wang_2024_CVPR, Chen_2024_CVPR} accelerate training by storing intermediate features in GPU buffers and jointly optimizing multiple views. However, directly applying such strategies to large-scale outdoor LiDAR localization is challenging. Outdoor localization benchmarks~\cite{Dan_2020_ICRA, Li_2023_CVPR} typically cover areas of up to 2 km$^2$ and contain approximately 150K training samples. Even storing sampled features, \emph{e.g.}, 1024 points with 512-dimensional features per scan, would require more than 150 GB of GPU memory under half-precision storage, making buffer-based strategies impractical. For large-scale outdoor LiDAR localization, we identify two key challenges in efficient scene-specific training: (1) extensive spatial coverage introduces many geometrically similar regions that increase regression ambiguity, (2) while the large data volume leads to substantial computational and storage costs.

In this paper, we integrate the SULID-pretrained sensor-robust backbone into the original LightLoc framework and develop LightLoc++. 
LightLoc++ retains two efficient training strategies from LightLoc~\cite{Li_2025_CVPR} to accelerate learning in large-scale outdoor scenes. Sample classification guidance (SCG) trains an auxiliary classification branch within a few minutes and uses the resulting probability distribution to guide scene coordinate regression, reducing ambiguity in geometrically similar regions. Redundant sample downsampling (RSD) identifies well-learned frames using the variance of the median loss and removes redundant samples, reducing computational cost without compromising localization accuracy. 
We evaluate LightLoc++ on multiple outdoor LiDAR localization benchmarks, including Oxford/QEOxford, Boreas, MulRan, and HeRCULES, with an overview shown in Fig.~\ref{fig:1}. We further conduct cross-sensor generalization experiments on HeLiPR~\cite{jung_2024_helipr}. The results demonstrate that LightLoc++ improves backbone generalization while preserving the fast scene-specific training capability of LightLoc.

Our contributions can be summarized as follows:
\begin{itemize}[leftmargin=2em]

\item LightLoc++, an efficient and sensor-robust outdoor LiDAR localization method, achieves state-of-the-art performance across multiple benchmarks collected with diverse LiDAR sensors. It can learn a new large-scale outdoor scene in just \textbf{1 hour}, while previous state-of-the-art SCR methods require about 2 days of training to achieve comparable accuracy.

\item SULID is introduced for sensor-robust representation learning. It provides synchronized observations from three representative rotating LiDARs, near-panoramic cross-sensor shared observations, and diverse large-scale urban scenes, making it a valuable resource for future research on sensor-robust perception and localization.

\item A sensor-robust LiDAR backbone is pretrained on SULID through cross-sensor consistency learning. In direct cross-sensor testing on HeLiPR, where the regressor is trained on Ouster and directly evaluated on Aeva and Velodyne without fine-tuning, the pretrained backbone shows substantially stronger generalization than existing methods.

\item We integrate two techniques, SCG and RSD, to address the challenges of large coverage and vast amounts of data in large-scale outdoor LiDAR localization, further improving the efficiency of new-scene learning.

\end{itemize}

This paper substantially extends our previous conference version, LightLoc (CVPR 2025)~\cite{Li_2025_CVPR}. Compared with the conference version, this paper expands the original focus on efficient scene-specific training toward sensor-robust representation learning for outdoor LiDAR localization. (1) We introduce SULID, a synchronized urban multi-LiDAR dataset that provides extensive cross-sensor observations from three representative rotating LiDAR sensors in diverse large-scale urban environments. (2) We pretrain a sensor-robust LiDAR backbone on SULID through cross-sensor consistency learning, improving the generalization ability of the scene-agnostic backbone under different LiDAR configurations. (3) We integrate the SULID-pretrained backbone into the LightLoc framework, resulting in LightLoc++ while preserving the efficient scene-specific training capability of the original method. (4) We provide substantially expanded experiments, including additional benchmarks, direct cross-sensor generalization evaluation, and detailed analyses of the learned representations. 
\section{Related Work}
\subsection{Map-based Localization}
Map-based LiDAR localization methods estimate poses by matching query point clouds to a pre-built 3D map~\cite{Yu_2021_ISPRS, Xia_2021_CVPR, Xia_2023_ICCV, Wang_2024_AAAI, Komorowski_2021_WACV, Uy_2018_CVPR, Luo_2023_ICCV, Wang_2019_ICCV, Choy_2019_CVPR, Qin_2022_CVPR, Ao_2023_CVPR, Zhang_2023_CVPR, Jin_2024_CVPR, liu2023regformer,luo2025bevplace++, qin2023geotransformer, wang2023roreg, yang2024mac}. 
These methods typically achieve high localization accuracy by leveraging explicit geometric information. 
However, they require storing and accessing large-scale maps, resulting in high storage and communication overhead, which limits their scalability in large-scale applications.

\subsection{Regression-based Localization}
To overcome the limitations of map-based methods, regression-based approaches have been proposed and have recently attracted significant attention. 
These methods can be broadly categorized into Absolute Pose Regression (APR) and Scene Coordinate Regression (SCR).

\noindent{\textbf{Absolute Pose Regression.}}
APR directly regresses the global 6-DoF pose of the sensor in an end-to-end manner~\cite{Kendall_2015_ICCV, Brahmbhatt_2018_CVPR, Wang_2020_AAAI, Yu_2022_PR, Wang_2023_AAAI, Chen_2024_CVPRB, Chen_2024_CVPR, Chen_2022_ECCV, shavit2023coarse}.
Early LiDAR APR methods, such as PointLoc~\cite{Wang_2022_Sensors}, adopt point-cloud backbones and attention mechanisms for pose regression.
Subsequent works~\cite{Yu_2022_PR, Wang_2023_CVPR} improve feature learning through enhanced backbones and multi-modal representations.
To further improve accuracy, some methods~\cite{Yu_2022_TITS, Yu_2023_TITS, Li_2024_CVPR} incorporate temporal information from multi-frame point clouds.
DiffLoc~\cite{Li_2024_CVPR, wang2025bevdiffloc} formulates pose estimation as a conditional generation problem and achieves strong APR performance, but introduces additional computational complexity and longer training times.

\noindent{\textbf{Scene Coordinate Regression.}}
In contrast to APR, SCR~\cite{Li_2023_CVPR, Yang_2024_CVPR,Yang_2025_ICCV,bruns2025ace,liu2025gs,wang2024hscnet++} predicts point-to-world correspondences and estimates poses via RANSAC~\cite{Fischler_1981_Commun}, explicitly leveraging scene geometry~\cite{Li_2023_CVPR}, achieving higher accuracy than APR. 

SGLoc~\cite{Li_2023_CVPR} decouples localization into correspondence prediction and pose estimation, achieving improved robustness and approximately 1.5-meter accuracy on QEOxford. 
LiSA~\cite{Yang_2024_CVPR} further enhances performance by incorporating semantic information, reaching an accuracy of 0.95 meters. 
RALoc~\cite{Yang_2025_ICCV} introduces additional regularization to improve correspondence learning and robustness under challenging conditions, further advancing SCR-based localization.
Despite their superior accuracy, SCR-based methods typically require substantially longer training times, often exceeding two days per large-scale outdoor scene.

\subsection{Efficient Training in Regression-based Localization}
Regression-based localization methods encode scene information into network parameters and avoid explicit map matching during inference. However, they typically require scene-specific training, which can take hours or even days for each new environment. Since training must be repeated independently for every scene, improving training efficiency is critical for practical deployment.

Recent works~\cite{Yu_2022_PR,Goswami_2025_WACV,Li_2025_CVPR} improve training efficiency by decoupling representation learning from scene-specific optimization. In this paradigm, a scene-agnostic backbone is pretrained and reused across scenes, while only lightweight scene-specific components are optimized for each new environment. Early work~\cite{Yu_2022_PR} demonstrates the feasibility of reusing pretrained backbones across scenes on datasets such as Oxford~\cite{Dan_2020_ICRA} and vReLoc~\cite{Wang_2022_Sensors}. FlashMix~\cite{Goswami_2025_WACV} pretrains a backbone with place recognition objectives on large-scale datasets such as MulRan~\cite{kim2020mulran} and Apollo SouthBay~\cite{lu_2019_l3}, and adapts it to new scenes by training only regression heads. LightLoc~\cite{Li_2025_CVPR} improves the efficiency of SCR by pretraining a scene-agnostic backbone on nuScenes~\cite{Caesar_2020_CVPR} using a multi-scene parallel optimization strategy~\cite{Brachmann_2023_CVPR}. GTR-Loc~\cite{yugtr_2025_NIPS} builds upon the pretrained backbone from LightLoc and incorporates geospatial text regularization to enhance localization performance.

Although these methods substantially reduce the cost of scene-specific training, their effectiveness relies heavily on the generalization ability of the pretrained backbone. Our analysis shows that when the target scene is collected with a LiDAR configuration different from that used during pretraining, the learned representation can become less reliable, leading to degraded localization performance, as shown in Fig.~\ref{fig:1}. This observation motivates us to improve the pretrained backbone itself through sensor-robust representation learning.

With a scene-agnostic backbone, efficient scene-specific training remains important for large-scale outdoor LiDAR localization. Outdoor benchmarks~\cite{Dan_2020_ICRA,Li_2023_CVPR} often cover large spatial areas and contain hundreds of thousands of training samples, leading to substantial computational and storage costs. Moreover, large-scale scenes contain many geometrically similar regions, which increases ambiguity in scene coordinate regression. To address these challenges, LightLoc introduces sample classification guidance and redundant sample downsampling to reduce regression ambiguity and computational redundancy. In this work, we integrate the sensor-robust backbone with these efficient training strategies, improving representation generalization while retaining the fast new-scene learning capability of LightLoc.

\begin{table*}[h!t]
\caption{\textbf{Comparison of LiDAR localization datasets.} 
LiDAR diversity refers to the use of sensors with different configurations (e.g., beam numbers and sampling densities), while cross-sensor consistency indicates aligned observations of the same scene across sensors. 
Scene types are categorized as \textbf{open} (sparse urban environments with weak structural constraints), 
\textbf{structured} (regular layouts with repeated geometric patterns), 
\textbf{dense} (cluttered environments with high structural density and strong occlusions), and 
\textbf{degenerate} (low-feature or geometrically ambiguous environments with limited observability).}
\centering

\begin{tabular}{l|c|cccccc}
\toprule
Datasets & Reference & Diverse LiDAR  & Cross-LiDAR Consistency & Diverse Scene & Scene Types  & Total Distance\\
\midrule
KITTI~\cite{geiger2012_CVPR}  & CVPR'12 & \textcolor{red}{\ding{55}}  & \textcolor{red}{\ding{55}} &  \textcolor{green}{\ding{51}} & Open  &  44 Km \\
NCLT~\cite{Carlone_2016_IJRR} & IJRR'16  & \textcolor{red}{\ding{55}} & \textcolor{red}{\ding{55}} & \textcolor{red}{\ding{55}} &  Structured  & 147 Km\\ 
Oxford~\cite{Dan_2020_ICRA} & ICRA'20 & \textcolor{red}{\ding{55}} & \textcolor{red}{\ding{55}} & \textcolor{red}{\ding{55}} &  Open  & 280 Km \\ 
MulRan~\cite{kim2020mulran} & ICRA'20 & \textcolor{red}{\ding{55}} & \textcolor{red}{\ding{55}} &  \textcolor{green}{\ding{51}} & Open, Structured  &  123 Km \\
Apollo-SouthBay~\cite{lu_2019_l3} & CVPR'19 & \textcolor{red}{\ding{55}} & \textcolor{red}{\ding{55}} &  \textcolor{green}{\ding{51}} &  Open  & 381 Km \\
nuScenes~\cite{Caesar_2020_CVPR} & CVPR'20 & \textcolor{red}{\ding{55}} & \textcolor{red}{\ding{55}} &  \textcolor{green}{\ding{51}} & Open  &  242 Km\\
Tiers~\cite{qingqing2022multi} & IROS'22 & \textcolor{green}{\ding{51}} & \textcolor{green}{\ding{51}} & \textcolor{green}{\ding{51}} & Structured  & 2.1 Km \\
BotanicGarden~\cite{liu2024botanicgarden} & RAL'23 &  \textcolor{green}{\ding{51}} &  \textcolor{green}{\ding{51}}  & \textcolor{red}{\ding{55}} & Degenerate   & 17 Km \\
Boreas~\cite{burnett2023boreas} & IJRR'23 & \textcolor{red}{\ding{55}} & \textcolor{red}{\ding{55}} & \textcolor{red}{\ding{55}} & Open   & 350 Km \\  
HeLiPR~\cite{jung_2024_helipr} & IJRR'24 &  \textcolor{green}{\ding{51}} & \textcolor{red}{\ding{55}} &  \textcolor{green}{\ding{51}} &  Open, Structured  & 164 Km \\ 
HeRCULES~\cite{kim2025hercules} & ICRA'25 & \textcolor{red}{\ding{55}} & \textcolor{red}{\ding{55}} &  \textcolor{green}{\ding{51}} &  Open, Structured  & 59 Km \\
GEODE~\cite{chen2026heterogeneous} & IJRR'26 & \textcolor{green}{\ding{51}} & \textcolor{red}{\ding{55}} &  \textcolor{green}{\ding{51}} & Degenerate   & 64 Km \\
\midrule
SULID & Ours &  \textcolor{green}{\ding{51}} &  \textcolor{green}{\ding{51}} &  \textcolor{green}{\ding{51}} &  Open, Structured, Dense  &  219 Km \\
\bottomrule
\end{tabular}

\label{tab:data}
\end{table*}

\subsection{Multi-LiDAR datasets}
Sensor-robust LiDAR representation learning requires cross-sensor observations of the same physical structures. Unlike mixed single-sensor datasets, synchronized multi-LiDAR datasets with large common observation areas provide direct correspondences for learning geometry-centric representations and evaluating cross-sensor localization. 

As summarized in Tab.~\ref{tab:data}, existing LiDAR localization datasets can be categorized into two groups according to LiDAR diversity. Widely used benchmarks such as KITTI~\cite{geiger2012_CVPR}, Oxford~\cite{Dan_2020_ICRA}, MulRan~\cite{kim2020mulran}, nuScenes~\cite{Caesar_2020_CVPR}, and Boreas~\cite{burnett2023boreas} provide valuable large-scale driving sequences for localization and perception. However, since each sequence is collected with a single sensor configuration, they cannot provide aligned observations of the same scene across different LiDAR sensors. This limits their use for studying cross-sensor consistency and sensor-robust representation learning.

Recent multi-LiDAR datasets, such as Tiers~\cite{qingqing2022multi}, HeLiPR~\cite{jung_2024_helipr}, BotanicGarden~\cite{liu2024botanicgarden}, and GEODE~\cite{chen2026heterogeneous}, introduce multiple LiDAR sensing configurations and enable evaluation under diverse sensing conditions. Nevertheless, as shown in Tab.~\ref{tab:data}, they still have different limitations. Tiers and BotanicGarden provide cross-LiDAR observations, but their scale is relatively limited and their scenes are mainly structured or degenerate. HeLiPR provides multiple LiDAR sensors and covers large-scale open and structured scenes, but its cross-LiDAR consistency is limited by the sensor mounting layout: the Ouster sensor is placed near the center of the platform and provides a 360$^\circ$ field of view, while the Aeva sensor is forward-facing and the Velodyne sensor is rear-mounted with partial occlusions. Due to differences in mounting positions, fields of view, and scanning patterns, LiDAR sensors in existing multi-LiDAR datasets often observe only partially overlapping regions, which limits the availability of shared physical structures for cross-sensor representation learning. GEODE focuses on heterogeneous sensing in degenerate environments, but its coverage of general large-scale urban scenes remains limited. As a result, existing multi-LiDAR datasets do not simultaneously provide diverse LiDAR configurations, extensive cross-LiDAR consistency, and diverse large-scale urban scene coverage.

To address these limitations, we construct SULID, a synchronized urban multi-LiDAR dataset designed for sensor-robust representation learning. SULID provides synchronized observations from representative 32-, 64-, and 128-beam rotating LiDARs, while the sensing platform is carefully designed to maximize the common observation area across sensors. In addition, SULID covers diverse urban scene types, including open, structured, and dense environments. These properties enable direct cross-sensor comparison and consistency learning, making SULID suitable for sensor-robust LiDAR representation learning and cross-sensor localization research.
\section{SULID}

\begin{figure*}[t]
  \centering
  \includegraphics[width=1\linewidth]{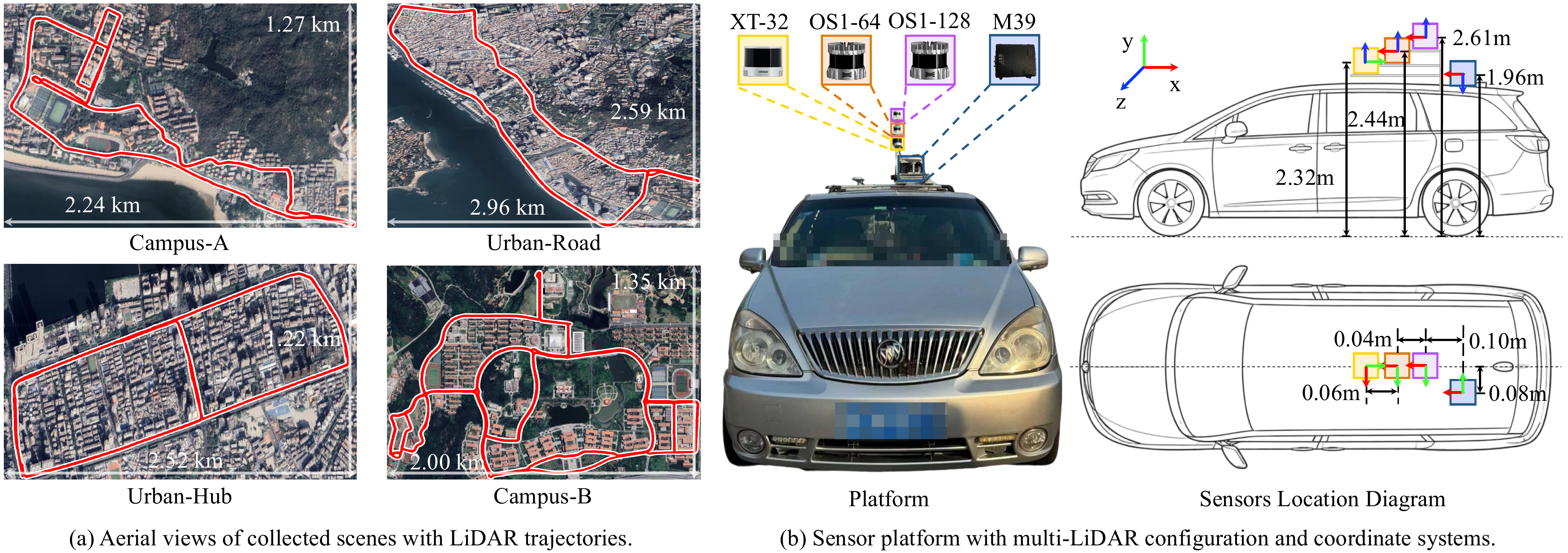}
   \caption{\textbf{Overview of the SULID dataset.} 
(a) Aerial views of the collected scenes, including \textbf{Campus-A}, \textbf{Campus-B}, \textbf{Urban-Road}, and \textbf{Urban-Hub}, with LiDAR trajectories overlaid, covering diverse urban environments. 
(b) The data collection platform is equipped with multiple LiDAR sensors of different configurations, along with their spatial arrangement and coordinate systems, enabling synchronized multi-LiDAR observations.
    }
   \label{fig:platform}
\end{figure*}

SULID is designed to support sensor-robust LiDAR representation learning in large-scale urban environments. Its design follows three main principles: diverse LiDAR configurations, extensive cross-sensor shared observations, and diverse urban scene coverage. To this end, we build a synchronized multi-LiDAR acquisition platform equipped with representative 32-, 64-, and 128-beam rotating LiDARs, and collect data across structured, open, and dense urban environments. The key characteristics of SULID are summarized in Tab.~\ref{tab:data}.

\subsection{Sensor Platform}
To provide diverse LiDAR observations of the same scene, we build a multi-LiDAR acquisition platform mounted on a Buick GL8, as shown in Fig.~\ref{fig:platform}~(b). The platform integrates three rotating LiDAR sensors with different configurations: Hesai XT-32, Ouster OS1-64, and Ouster OS1-128. As summarized in Tab.~\ref{tab:sensor}, these sensors differ in vertical resolution, field of view, and point density, covering representative 32-, 64-, and 128-beam LiDAR configurations. All LiDARs operate at 10 Hz. A high-precision GNSS/INS system (M39) is also installed to provide ego-motion estimates at 200 Hz.

\begin{figure}[t]
  \centering
  \includegraphics[width=1\columnwidth]{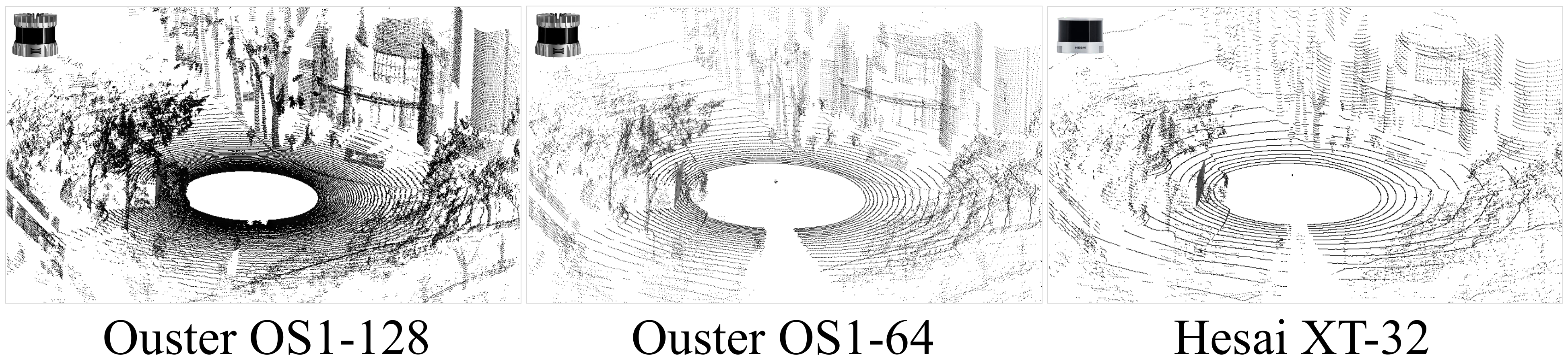}
   \caption{\textbf{Cross-sensor point-cloud observations in SULID.} Point clouds captured simultaneously at the same location by Ouster OS1-128, Ouster OS1-64, and Hesai XT-32. The sensors exhibit different sampling densities and spatial patterns while preserving largely overlapping scene structures, providing shared observations for cross-sensor representation learning.}
   \label{fig:pc_compare}
\end{figure}

All sensors are rigidly mounted on the vehicle roof to maintain fixed relative poses during data collection. Since different LiDAR configurations produce distinct point-cloud characteristics, the platform is carefully designed to reduce mutual occlusions and enlarge the common observation area across sensors. In particular, the LiDAR sensors are mounted at different heights and horizontally offset positions, as shown in Fig.~\ref{fig:platform}~(b), allowing them to observe largely overlapping scene regions while preserving their distinct LiDAR configurations. Fig.~\ref{fig:pc_compare} shows that synchronized point clouds captured at the same location exhibit different point densities and spatial distributions, while still preserving largely overlapping scene structures. This design provides rich cross-sensor observations of the same physical scene, forming a natural basis for learning sensor-robust representations.

\subsection{Synchronization and Calibration}
\noindent{\textbf{Synchronization.}}
To ensure temporal consistency, all LiDAR sensors are synchronized using Precision Time Protocol (PTP)~\cite{ieee1588}, with the GNSS/INS system providing a shared time reference. In addition, synchronized triggering is used to align the acquisition cycles of different LiDAR sensors. This setup enables point clouds from different LiDAR sensors to be captured at the same nominal timestamp, ensuring that synchronized scans correspond to the same underlying scene. Such temporal alignment is essential for constructing reliable cross-sensor correspondences and supporting cross-sensor consistency learning.

\begin{table}[!t]
\caption{\textbf{Specifications of multi-LiDAR sensors used in SULID.}}
\centering

\begin{tabular}{l|ccccc}
\toprule
Sensor & Resolution (V$\times$H) & FOV (H$\times$V) & Range  & FPS\\

\midrule

Ouster OS1-128 & 128$\times$2048  & 360$^{\circ}$$\times$45$^{\circ}$ & 120m & 10Hz\\

Ouster OS1-64 & 64$\times$1024  & 360$^{\circ}$$\times$45$^{\circ}$ & 120m &10Hz\\

Hesai XT-32 & 32$\times$1800  & 360$^{\circ}$$\times$31$^{\circ}$ & 120m &10Hz\\
\bottomrule
\end{tabular}
\label{tab:sensor}
\end{table}

\noindent{\textbf{Calibration.}}
For spatial alignment, we perform both LiDAR-to-LiDAR and LiDAR-to-GNSS/INS calibration. 
LiDAR-to-LiDAR calibration is conducted in scenes with rigid structures and distinctive geometric features, where the relative transformations between sensors are estimated using point cloud registration~\cite{segal2009generalized}.
For LiDAR-to-GNSS/INS calibration, we follow a standard calibration procedure similar to KITTI~\cite{geiger2012_CVPR}, estimating the rigid transformation between the reference LiDAR frame and the GNSS/INS frame by aligning LiDAR observations with ego-motion trajectories provided by the GNSS/INS system. The calibrated extrinsics are used to associate synchronized multi-LiDAR observations with ego poses and transform point clouds into a unified coordinate system.

\noindent{\textbf{Pose Reference.}}
Accurate pose references are obtained using the post-processing software provided with the GNSS/INS system (M39). The system records raw GNSS and INS measurements during data collection, and the post-processing pipeline fuses these measurements to generate stable and temporally consistent ego-motion trajectories. The resulting poses are used as supervision for training and evaluation.

\begin{figure*}[t]
  \centering
  \includegraphics[width=1\linewidth]{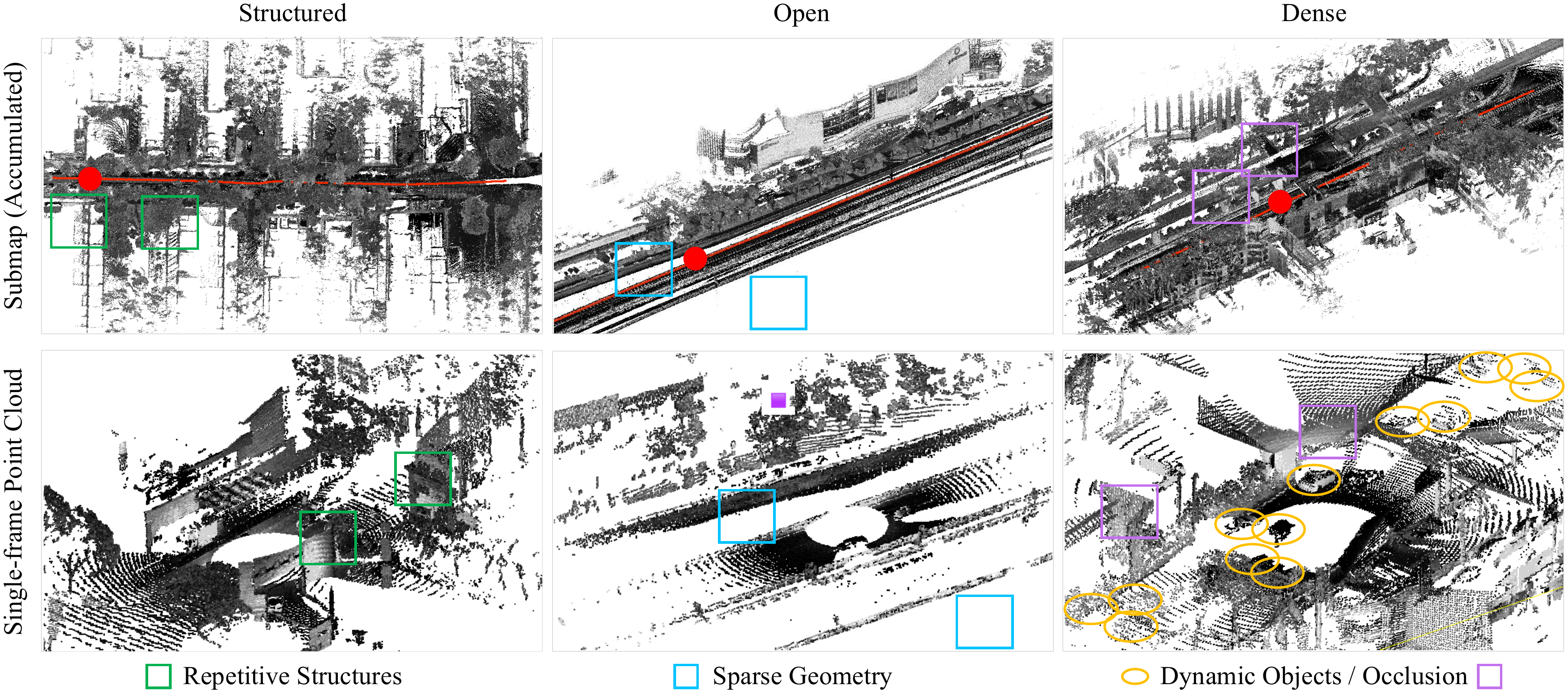}
   \caption{\textbf{Representative scene types in SULID.} 
Each column corresponds to a scene type, including structured, open, and dense environments. 
The top row shows accumulated point cloud maps (submaps), while the bottom row presents single-frame point clouds. 
Structured environments exhibit repetitive geometric patterns, open environments contain sparse structures with weak geometric constraints, and dense environments are characterized by strong occlusions and clutter. The \textbf{\textcolor{red}{red line}} indicates the driving trajectory. The \textbf{\textcolor{red}{red dot}} highlights the reference position.
    }
   \label{fig:scene}
\end{figure*}

\subsection{Scene Design and Data Collection}
To support sensor-robust representation learning under diverse urban conditions, SULID is collected across four urban regions: Campus-A, Campus-B, Urban-Road, and Urban-Hub, as shown in Fig.~\ref{fig:platform}~(a). These regions are organized into three representative scene types: structured, open, and dense environments. The aerial views and overlaid trajectories in Fig.~\ref{fig:platform}~(a) show the spatial coverage of each region, while Fig.~\ref{fig:scene} presents representative accumulated submaps and single-frame point clouds. These scene types introduce complementary challenges for LiDAR localization, including structural repetition, sparse geometric constraints, and dynamic occlusions.

\noindent{\textbf{Campus~(Structured Scenario).}}
The structured scenes are collected in campus areas, including Campus-A and Campus-B. These environments are characterized by regular building layouts, organized road networks, and repeated roadside elements, as shown in Fig.~\ref{fig:scene}. Such regular structures provide stable geometric cues, but they may also introduce local ambiguity because similar facades, repeated buildings, and regularly arranged objects can produce visually similar observations at different locations.

SULID includes two campus subsets with complementary spatial coverage. Campus-A contains 6 trajectories, each approximately 8.0 km, with largely overlapping routes. Campus-B contains 8 trajectories, each approximately 9.7 km, covering a broader spatial extent under similar structured layouts. These repeated traversals allow the evaluation of representation consistency under revisits, structural regularity, and repetitive geometric patterns.

\begin{figure}[t]
  \centering
  \includegraphics[width=1\columnwidth]{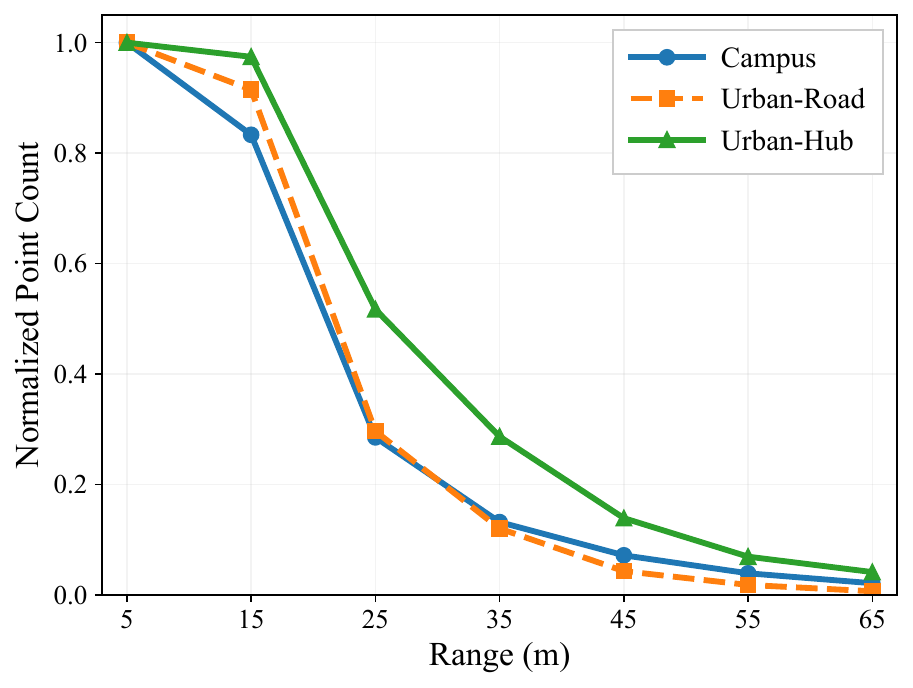}
\caption{\textbf{Range-dependent point distributions across different scene types.} Point counts are computed within distance bins using Ouster OS1-128 scans and normalized by the nearest-range bin. The structured, open, and dense scenes exhibit distinct distribution patterns, reflecting differences in geometric sparsity, structural density, and occlusion characteristics.}
   \label{fig:pc_distribution}
\end{figure}

\noindent{\textbf{Urban-Road~(Open Scenario).}}
The open scenes are collected in Urban-Road, which covers major urban roads and a cross-sea bridge. This environment is dominated by wide road surfaces, open surroundings, and relatively sparse nearby structures. Compared with structured scenes, open environments provide fewer distinctive geometric elements, making it more challenging to extract stable localization cues from single-frame LiDAR observations.

We collect 5 trajectories in Urban-Road, each approximately 8.8 km, with repeated traversals along similar routes. This scene type is designed to evaluate whether the learned representation can remain reliable when local geometric constraints are weak and informative structures are sparsely distributed.

\noindent{\textbf{Urban-Hub~(Dense Scenario).}}
The dense scenes are collected in Urban-Hub, a major urban transportation area with complex road layouts, dense roadside facilities, heavy traffic, and frequent occlusions. In contrast to open scenes, Urban-Hub contains abundant geometric details, but many observations are affected by dynamic objects, such as vehicles and pedestrians. This requires the learned representation to focus on stable scene geometry while being robust to clutter and objects.

SULID collects 5 trajectories in Urban-Hub, each approximately 9.8 km. The trajectories cover similar spatial regions under different traffic conditions and dynamic patterns, providing challenging data for evaluating localization robustness in dense and dynamic urban environments.

Overall, SULID contains 24 trajectories across multiple urban regions, covering approximately 219 km. To further quantify the geometric differences among scene types, we analyze the range-dependent point distributions using Ouster OS1-128, as shown in Fig.~\ref{fig:pc_distribution}. The three scene categories exhibit distinct distribution patterns, reflecting differences in structural regularity, geometric sparsity, and occlusion characteristics.

\begin{figure*}
  \centering
  \includegraphics[width=0.9\linewidth]{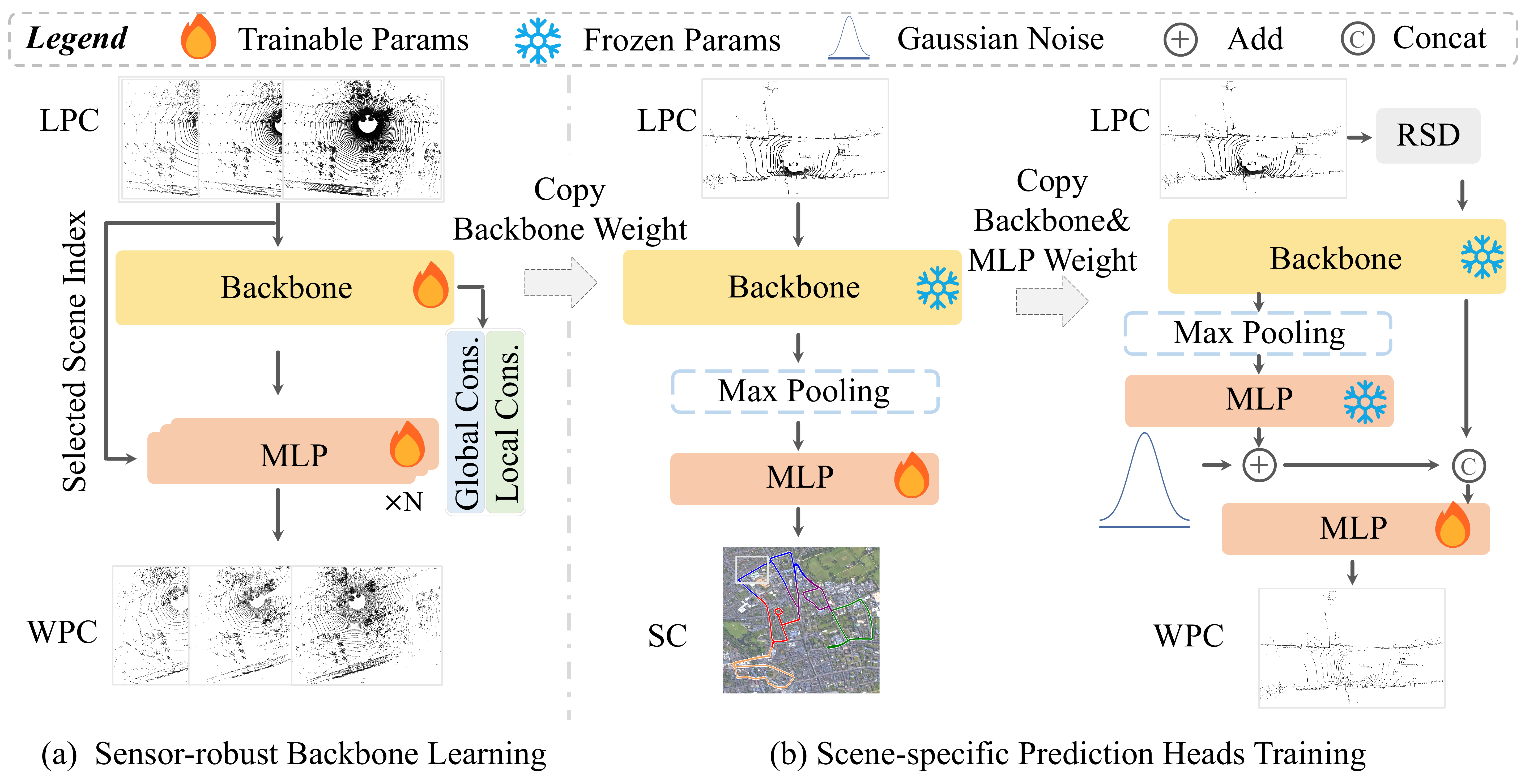}
\caption{
\textbf{Training pipeline of LightLoc++.}
(a) In sensor-robust backbone learning, synchronized scans are transformed into a unified LiDAR coordinate system and fed into a shared backbone with multi-scene regression heads. Local and global consistency losses are imposed across synchronized LiDAR observations to learn sensor-robust representations.
(b) In new-scene learning, the pretrained backbone is frozen, while lightweight scene-specific heads are trained with \textbf{SCG} and \textbf{RSD} for efficient SCR.
\textbf{LPC} and \textbf{WPC} denote local point coordinates in the unified LiDAR coordinate system and world point coordinates, respectively. \textbf{SC} denotes sample classification.
}
  \label{fig:pipline}
\end{figure*}

\subsection{Dataset Summary}
As summarized in Tab.~\ref{tab:data}, SULID provides synchronized multi-LiDAR observations across diverse large-scale urban environments. The sensor platform and sensor specifications are shown in Fig.~\ref{fig:platform} and Tab.~\ref{tab:sensor}, respectively, demonstrating that SULID covers representative 32-, 64-, and 128-beam LiDAR configurations. The extensive cross-sensor shared observations are illustrated in Fig.~\ref{fig:pc_compare}, while the structured, open, and dense urban scenes are shown in Fig.~\ref{fig:scene}. Overall, SULID contains 24 trajectories covering approximately 219 km. These properties provide the data foundation for the sensor-robust representation learning framework introduced in the next section.
\section{LightLoc++}

\subsection{Overview}\label{Sec.4.1}
We now present LightLoc++, as illustrated in Fig.~\ref{fig:pipline}. The proposed framework follows the decoupled training paradigm introduced in previous works~\cite{Li_2025_CVPR,Brachmann_2023_CVPR, Li_2025_CVPR}, where scene-agnostic representation learning and scene-specific prediction head training are separated.

LightLoc++ consists of two main components. The first component is sensor-robust backbone learning, which leverages synchronized observations from multiple LiDAR sensors to learn representations that remain effective across different LiDAR configurations. By enforcing cross-sensor consistency during pretraining, the backbone is encouraged to capture stable scene geometry while reducing sensitivity to sensor-specific configurations. The second component is scene-specific prediction head training, where the pretrained backbone is frozen and lightweight scene-specific heads are optimized for new scenes.

\begin{figure*}
  \centering
  \includegraphics[width=1\linewidth]{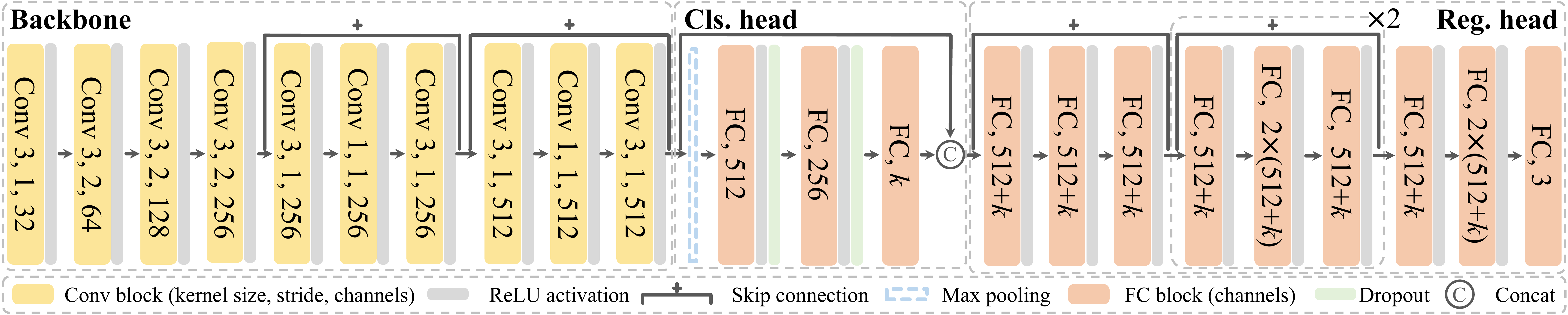}
\caption{\textbf{Network architectures used in LightLoc++.} 
LightLoc++ uses the same backbone and prediction-head architectures as LightLoc~\cite{Li_2025_CVPR}. 
\textbf{Cls. head} and \textbf{Reg. head} denote the sample classification head and the scene coordinate regression head, respectively. 
The backbone extracts scene-agnostic point features, the classification head provides spatial guidance for SCG, and the regression head predicts scene coordinates for pose estimation. 
Here, $k$ denotes the number of spatial clusters used for sample classification in SCG.}
  \label{fig:structure}
\end{figure*}

Compared with LightLoc, the key extension of LightLoc++ lies in the backbone learning stage. Instead of pretraining the backbone on a single-LiDAR dataset, LightLoc++ uses synchronized multi-LiDAR observations and cross-sensor consistency learning to improve backbone generalization under different LiDAR configurations. The network architectures, including the backbone and prediction heads, remain the same as LightLoc~\cite{Li_2025_CVPR}, as illustrated in Fig.~\ref{fig:structure}. Meanwhile, the scene-specific training pipeline is retained, preserving the fast new-scene learning capability of LightLoc.

\subsection{Revisiting LightLoc}\label{Sec.4.2}
We first briefly revisit LightLoc~\cite{Li_2025_CVPR}. LightLoc adopts a decoupled training paradigm that separates backbone learning from scene-specific prediction head training.

For the network architecture, LightLoc adopts the backbone of SGLoc~\cite{Li_2023_CVPR} with a reduced feature dimension and fewer residual layers to reduce the number of parameters and shorten the training time. In the backbone training stage, following ACE~\cite{Brachmann_2023_CVPR}, LightLoc trains $N$ regression heads in parallel, each corresponding to one of $N$ training scenes. Specifically, given an input point cloud ${\cal P} = \{p_i\}_{i=1}^M$, where $M$ is the number of input points and $p_i \in \mathbb{R}^{3}$ denotes a local point coordinate (LPC) in the LiDAR coordinate system, the backbone network $f_\theta$ is used to extract point-wise features. Given the scene index $\mathrm{idx}$, the corresponding regression head $h_{\mathrm{idx}}$ is selected to predict the world point coordinates (WPC):
$h_{\mathrm{idx}}\left(f_{\theta}\left({\cal P}\right)\right) = \{p_i'\}_{i=1}^{M'}$,
where $M'$ denotes the number of output points after feature extraction. In LightLoc, $M'=M/8$ due to the downsampling in the backbone. The training objective minimizes the $L_1$ distance between the predicted WPC and the ground-truth WPC:
\begin{equation}
\mathcal{L}_{\text{reg}} =
\frac{1}{M'}
\sum_{i=1}^{M'} \left\| p_i' - p_i^* \right\|_1 .
\label{eq_reg}
\end{equation}

This multi-scene parallel training strategy encourages the backbone to learn features that generalize across diverse scenes. In practice, LightLoc partitions nuScenes~\cite{Caesar_2020_CVPR}, including both the trainval and test splits, into 18 spatial training scenes using k-means clustering over trajectory locations. The partitioned data contain approximately 350K training samples, with about 19.5K samples per partition on average. Each partition is assigned to an independent regression head during multi-scene parallel backbone training.

In addition to freezing the backbone for new-scene learning, LightLoc introduces two complementary strategies to address the training challenges caused by large spatial coverage and a large number of training samples. First, sample classification guidance (SCG) uses an auxiliary spatial classification task to provide global guidance for scene coordinate regression, reducing ambiguity in geometrically similar regions. Second, redundant sample downsampling (RSD) dynamically filters out well-learned samples according to loss statistics, reducing computational redundancy during training.

During inference, similar to SGLoc~\cite{Li_2023_CVPR}, LightLoc takes LPC as input and directly predicts their corresponding WPC. The final 6-DoF pose is then estimated from these correspondences using RANSAC~\cite{Fischler_1981_Commun}.

\subsection{Sensor-Robust Backbone Training}\label{Sec.4.3}
LightLoc achieves strong performance on Oxford~\cite{Dan_2020_ICRA} and QEOxford~\cite{Li_2023_CVPR}, where the LiDAR configuration is relatively close to that of the pretraining data. However, on datasets collected with different LiDAR sensors, such as Boreas~\cite{burnett2023boreas}, MulRan~\cite{kim2025hercules}, and HeRCULES~\cite{kim2025hercules}, the performance gap compared with conventional SCR methods becomes much larger. These results indicate that the effectiveness of decoupled SCR methods largely depends on the generalization ability of the pretrained backbone, which remains limited under LiDAR sensor variations.

We argue that a robust pretrained backbone should capture geometric properties of the scene that remain stable across different LiDAR configurations. Therefore, we leverage the synchronized multi-LiDAR observations provided by SULID to learn sensor-robust representations. Following LightLoc, we divide SULID into multiple spatial training scenes and attach a dedicated regression head to each scene.

Given a synchronized timestamp, we use the calibrated LiDAR-to-LiDAR extrinsics to transform point clouds from different sensors into a unified LiDAR coordinate system. The resulting synchronized point cloud from sensor $s$ is denoted as $\mathcal{P}^{s}$, where $s \in \mathcal{S}$ and $\mathcal{S}=\{\mathrm{O128}, \mathrm{O64}, \mathrm{H32}\}$ denotes the Ouster OS1-128, Ouster OS1-64, and Hesai XT-32 sensors, respectively. Each point cloud $\mathcal{P}^{s}$ is then fed into the shared backbone $f_\theta$ to extract point-wise features:
\begin{equation}
\mathcal{F}^{s}=f_\theta\left(\mathcal{P}^{s}\right), \quad s\in\mathcal{S}.
\label{eq_f}
\end{equation}

Inspired by contrastive representation learning~\cite{xie2020pointcontrast,nunes2022segcontrast}, we impose cross-sensor consistency constraints on the extracted features $\mathcal{F}^{s}$, including global scan-level consistency and local point-level consistency.

\noindent{\textbf{Global Consistency.}}
The synchronized point clouds from different LiDAR sensors correspond to the same timestamp and observe the same geometric properties. Therefore, we first align their scene-level representations.

Let $g(\cdot)$ denote a permutation-invariant pooling operator that maps a sparse feature set to a unit-norm global descriptor. For sensor $s$, the global descriptor is defined as $g^{s}=g(\mathcal{F}^{s})$. We then compute the cross-sensor consensus descriptor as $\bar{g}=\tfrac{1}{|\mathcal{S}|}\sum_{s\in\mathcal{S}}g^{s}$. We enforce global consistency by minimizing the deviation of each sensor-specific descriptor from the consensus:
\begin{equation}
\mathcal{L}_{\text{global}}
=
\frac{1}{|\mathcal{S}|}
\sum_{s\in\mathcal{S}}
\left\|
g^{s} - \bar{g}
\right\|_2 .
\end{equation}

\noindent{\textbf{Local Consistency.}}
Since synchronized point clouds from different LiDAR sensors are transformed into a common LiDAR coordinate frame, their point-level features can be compared according to their 3D positions. We therefore perform geometric nearest-neighbor matching across sensors to identify reliable shared points. A point is retained only when its nearest-neighbor distance across sensors is smaller than a predefined threshold, yielding a set of reliable shared points denoted as $\mathcal{V}$.
For each shared point $v\in\mathcal{V}$, let $F_v^{s}$ denote the local feature extracted from sensor $s$, and define the cross-sensor consensus feature as $\bar{F}_v=\frac{1}{|\mathcal{S}|}\sum_{s\in\mathcal{S}}F_v^{s}$. Then, the local consistency loss is defined as follows:
\begin{equation}
\mathcal{L}_{\text{local}}
=
\frac{1}{|\mathcal{V}||\mathcal{S}|}
\sum_{v\in\mathcal{V}}
\sum_{s\in\mathcal{S}}
\left\|
F_v^{s} - \bar{F}_v
\right\|_2 .
\end{equation}

Defined on reliable cross-sensor observations, the local consistency term provides per-location supervision for learning sensor-invariant features on shared scene structures.

The backbone loss combines the two terms:
\begin{equation}
\mathcal{L}_{\text{cons}}
=
\mathcal{L}_{\text{local}}
+
\mathcal{L}_{\text{global}},
\label{eq_cons}
\end{equation}

\noindent Despite its simple formulation, $\mathcal{L}_{\text{cons}}$ is enabled by the tri-sensor synchronization and shared coordinate frame provided by SULID, which allows reliable cross-sensor feature alignment at both global and local levels.

\subsection{Scene-specific Prediction Heads Training}\label{Sec.4.4}

As shown in Fig.~\ref{fig:pipline}, after obtaining the pretrained backbone, we freeze its parameters for new scenes, 
training only scene-specific prediction heads.  
However, two challenges that remain for accelerating SCR training in large-scale outdoor scenes: the extensive coverage area, which complicates regression learning, and the vast amount of data, which imposes a significant computational burden. To address these issues, we adopt SCG and RSD techniques.

\noindent{\textbf{SCG.}}
SCG is designed to generate a sample-wise probability distribution feature with only a few minutes of auxiliary training. The generated feature is then used to guide SCR training, reducing the training time required for new scenes. Its network structure is illustrated in Fig.~\ref{fig:structure}, where the auxiliary training objective is formulated as a sample classification task.

We first generate classification labels from the training trajectory. Specifically, K-Means is applied to partition the training positions into $k$ spatial clusters, and the cluster index of each sample is used as its pseudo classification label. This label generation process requires no manual annotation and provides a coarse spatial partition of the target scene.

Subsequently, we train the classification network. 
For the input point cloud $\mathcal{P}$, 
we use the pretrained backbone $f_\theta$ to extract features $f_\theta(\mathcal{P})$. 
A global max-pooling operation is then applied to aggregate the point-wise features into a global feature vector, which is fed into an MLP classifier to predict the probability distribution over the $k$ spatial clusters. The classifier is trained using the cross-entropy loss with label smoothing $(\epsilon=0.1)$~\cite{Muller_2019_NIPS}:
\begin{equation}
{\cal L}_{\text{cls}}= -\sum_{i=1}^{k_1} \left(l^{*}_i\left(1-\epsilon\right) + \frac{\epsilon}{k_1}\right)log{\left(l^{'}_i\right)},
\label{eq_cls}
\end{equation}

\noindent where $l^{'}_i$ and $l^{*}_i$ denote the predicted probability and ground-truth label of class $i$, respectively.

After training, the predicted probability distribution is used as the SCG feature to provide coarse spatial guidance for SCR. As shown in the right part of Fig.~\ref{fig:pipline}~(b), following GLACE~\cite{Wang_2024_CVPR,jiang2025r}, we further perturb the SCG feature with Gaussian noise of standard deviation $\sigma=0.1$ and normalize it back to the unit sphere. The resulting guidance feature is incorporated into the SCR framework, providing scene-level spatial priors that facilitate faster convergence and improve the robustness of regression learning.

\begin{algorithm}[t]
    \SetKwInput{KwInput}{Input}
    \SetKwInput{KwOutput}{Return}
    \DontPrintSemicolon
    \SetAlgoLined
    \SetNoFillComment
    \caption{{\samplfull}.}
    \label{alg:1}
    \SetAlgoVlined
    \KwInput{Training set $\cal T$, downsampling ratio $r_d$, start ratio $r_{st}$, stop ratio $r_{sp}$, training epochs $E$, length of sliding window $W$.}
    First downsampling epochs $E_1\gets\lceil E\times r_{st}\rceil$. \\
    Second downsampling epochs $E_2\gets\lfloor E\times \frac{r_{st}+r_{sp}}{2}\rfloor$.\\
    Stop downsampling epochs $E_s \gets \lfloor E\times r_{sp} \rfloor$. \\
    $\cal T^{'} \gets \cal T$. \\
    \For {$e=0, \ldots, E-1$}{
        $\triangleright$ In parallel on GPUs.\\
		Sample a batch $\cal{B}\sim \cal{T^{'}}$. \\
        \For{$({\cal P}_i,{\cal P}_i^{*}) \in \cal{B}$}{
            Compute the median point-wise $L_1$ loss as ${\cal L}_{m}^{i}$. \\
            \If{$e \in \left[ E_j, E_j+W\right)$}
            {\tcp{$j \in \left[1,2 \right]$} 
            Record ${\cal L}_{m}^{i}$ within $W$.
            }
        }
        \If{$e = E_j + S$}{
        Compute the loss variance $\sigma_i^2$ of ${\cal L}_{m}^{i}$ within $W$.\\
        Sort ${\cal T}^{'}$ in descending order of $\cal V$.\\
        ${\cal T}^{'} \gets$ front $(1-r_d)\times\left|{\cal T}^{'}\right|$ samples.
        }
        \If{$e = E_s$}{
        ${\cal T}^{'} \gets \cal T$.
        }
    }
    \KwOutput{${\cal T}^{'}$.}
\end{algorithm}

\noindent{\textbf{RSD.}}
Large-scale outdoor LiDAR datasets usually contain substantial redundancy due to the long sensing range (up to 100m) and high acquisition frequency (10Hz). To reduce unnecessary computation during scene-specific SCR training, we introduce the RSD strategy, as shown in Alg.~\ref{alg:1}.

RSD performs hierarchical sample downsampling by partitioning the training process into multiple stages, 
controlled by predefined epochs and a downsampling ratio. 
Specifically, we define $E_1$ epochs for the initial stage, $E_2$ for the intermediate stage, and $E_s$ as the final epoch, 
together with a downsampling ratio $r_d$. Below, we describe each stage.

In the first stage, we optimize SCR using the full training set $\mathcal{T}$ for $E_1$ epochs. For each sample, we compute the median point-level $L_1$ loss, denoted as $\mathcal{L}_m$, where the median is used for its robustness to outliers.

In the second stage, we assess the convergence of each sample by calculating the variance of $\mathcal{L}_m$ within a sliding window of size $W$. We denote the loss variance of sample $i$ as $\sigma_i^2$. At epoch $E_1+W$, we sort the samples in $\mathcal{T}$ by $\sigma_i^2$ in descending order and retain the top $(1-r_d)|\mathcal{T}|$ samples as the reduced set $\mathcal{T}'$. High-variance samples are prioritized because they indicate slower convergence and therefore require more training focus.

In the third stage, we repeat the same procedure on $\mathcal{T}'$. Specifically, we compute the median loss and its variance within the sliding window of size $W$, and further retain the top $(1-r_d)|\mathcal{T}'|$ samples according to $\sigma_i^2$. As a result, the training set is reduced to $(1-r_d)^2|\mathcal{T}|$ samples.

In the final stage, we reintroduce the full training set $\mathcal{T}$ from epoch $E_s$ onward to ensure that all samples are sufficiently optimized before convergence.

With RSD, the network can exclude well-converged samples during intermediate training stages by simply computing loss variance within a sliding window. This strategy introduces minimal overhead, accelerates SCR training, and preserves localization accuracy.

\subsection{Loss Function.}\label{Sec.4.5}

The training objective of LightLoc++ consists of two stages: sensor-robust backbone pretraining and scene-specific prediction head optimization.

For backbone pretraining, we jointly optimize the scene coordinate regression loss $\mathcal{L}_{\text{reg}}$ and the cross-sensor consistency loss $\mathcal{L}_{\text{cons}}$. The regression loss preserves the task-specific localization capability of the backbone, while the consistency loss encourages synchronized LiDAR scans from different sensors to produce sensor-invariant representations. The backbone training objective is defined as:
\begin{equation}
\mathcal{L}_{\text{backbone}}
=
\mathcal{L}_{\text{reg}}
+
\lambda \mathcal{L}_{\text{cons}},
\label{eq_backbone}
\end{equation}

\noindent where $\lambda$ is a balancing weight that controls the contribution of the cross-sensor consistency loss during backbone pretraining.

For scene-specific prediction head training, we optimize two objectives: the auxiliary classification loss $\mathcal{L}_{\text{cls}}$ and the scene coordinate regression loss $\mathcal{L}_{\text{reg}}$. Specifically, $\mathcal{L}_{\text{cls}}$ is used to train the SCG branch with cross-entropy loss and a label smoothing factor of $\epsilon=0.1$, as defined in Eq.~\ref{eq_cls}. The learned sample probability distribution is then used as spatial guidance for SCR, where the regression head is optimized using $\mathcal{L}_{\text{reg}}$ defined in Eq.~\ref{eq_reg}.
\section{Experiments}
\subsection{Datasets}
We evaluate LightLoc++ under diverse LiDAR configurations and large-scale outdoor environments. Specifically, localization benchmarking is conducted on five datasets: Oxford Radar RobotCar (Oxford)~\cite{Dan_2020_ICRA}, QEOxford~\cite{Li_2023_CVPR}, MulRan~\cite{kim2020mulran}, Boreas~\cite{burnett2023boreas}, and HeRCULES~\cite{kim2025hercules}. These datasets cover different LiDAR sensors, acquisition platforms, urban layouts, and environmental conditions, enabling a comprehensive evaluation of localization accuracy and cross-dataset robustness. In addition, we use HeLiPR~\cite{jung_2024_helipr} for cross-sensor generalization evaluation, where models trained with other LiDAR sensors are tested on another to assess sensor robustness.

\noindent{\textbf{Oxford Radar RobotCar (Oxford)}} is collected using an autonomous-capable Nissan LEAF platform for urban localization. Each trajectory spans approximately 10 km and covers an area of about 2 km$^2$. Point clouds are captured by the Velodyne HDL-32E LiDAR sensor, and ground-truth poses are provided by a GPS/INS system. The dataset contains repeated traversals under diverse weather, illumination, and traffic conditions, making it a widely used benchmark for urban localization.

\noindent{\textbf{QEOxford}} is a quality-enhanced version of Oxford, where GPS/INS pose errors are reduced through trajectory alignment. The improved pose quality has been shown to benefit scene coordinate regression-based localization~\cite{Li_2023_CVPR}. For both Oxford and QEOxford, we follow SGLoc~\cite{Li_2023_CVPR}. The sequences 11-14-02-26, 14-12-05-52, 14-14-48-55, and 18-15-20-12 are used for training, while 15-13-06-37, 17-13-26-39, 17-14-03-00, and 18-14-14-42 are used for testing.

\noindent{\textbf{Boreas}} is a multi-season autonomous driving dataset collected with a Velodyne Alpha-Prime 128-beam LiDAR. Each trajectory spans approximately 9 km, with repeated traversals under different seasonal and weather conditions. Its high-resolution LiDAR measurements and significant environmental variations make it challenging for localization. We use the sequences 2020-11-26, 2021-03-23, 2021-03-30, and 2021-04-13 for training, and 2020-12-01, 2021-02-02, 2021-03-02, and 2021-04-29 for testing.

\noindent{\textbf{MulRan}} was collected using a vehicle-mounted platform equipped with an Ouster OS1-64 LiDAR sensor, covering multiple urban environments. Following FlashMix~\cite{Goswami_2025_WACV}, we use the DCC sequence, which features dense urban structures and has an average trajectory length of approximately 4.8 km. Specifically, DCC01 and DCC02 are used for training, and DCC03 is used for testing.

\noindent{\textbf{HeRCULES}} is a heterogeneous sensing dataset that includes LiDAR measurements acquired with an Aeva Aeries II sensor, whose sensing characteristics differ from conventional LiDARs. 
In this work, we use the Library and Sports Complex sequences.
The Library sequence spans approximately 1.6 km and features a narrow campus road with curves and elevation changes, while the Sports Complex sequence spans approximately 1.4 km, covering open areas with parking lots and roads of varying slopes.

In the Sports scene, Sports01 and Sports02 are used for training, and Sports03 is used for testing. In the Library scene, Library01 and Library02 are used for training, and Library03 is used for testing. 

\noindent{\textbf{HeLiPR}} is a heterogeneous LiDAR dataset used for cross-sensor generalization evaluation. We use the DCC trajectories, with an average length of approximately 5.2 km, where DCC04 and DCC06 are used for training and DCC05 for testing. The model is trained only on Ouster OS2-128 data and evaluated on Ouster OS2-128, Aeva Aeries II, and Velodyne VLP-16 data, allowing us to assess both same-sensor performance and cross-sensor robustness.

\begin{table*}[t]
	\centering
\caption{\textbf{Quantitative results on the QEOxford dataset.} Mean ATE (m)  /  ARE ($^{\circ}$) are reported. . Best results in \textbf{bold}, second best results \underline{underlined}. Training time is also included.}
\centering
\begin{tabular}{c|@{}l|c|c|cccc|c}
\toprule

& \quad Methods & Reference & Training Time  & {15-13-06-37} & {17-13-26-39} &  {17-14-03-00} & {18-14-14-42} & Average\\


%
\midrule
& \quad PointLoc~\cite{Wang_2022_Sensors}  & Sens. J.'22 &  125 hrs.  & 10.75 / 2.36 & 11.07 / 2.21 & 11.53 / 1.92 & 9.82 / 2.07 & 10.79 / 2.14 \\
& \quad PosePN~\cite{Yu_2022_PR}  & PR'22 &  5 hrs.  & 9.47 / 2.80 & 12.98 / 2.35 &  8.64 / 2.19 & 6.26 / 1.64 & 9.34 / 2.25 \\
& \quad PosePN++~\cite{Yu_2022_PR}  & PR'22 &  11 hrs.  & 4.54 / 1.83 & 6.44 / 1.78 & 4.89 / 1.55 & 4.64 / 1.61 & 5.13 / 1.69 \\
& \quad STCLoc~\cite{Yu_2022_TITS}   & TITS'23 &  20 hrs.  & 5.14 / 1.27 & 6.12 / 1.21 & 5.32 / 1.08 & 4.76 / 1.19 & 5.34 / 1.19 \\
& \quad NIDALoc~\cite{Yu_2023_TITS} & TITS'24 &  20 hrs.  & 3.71 / 1.50 & 5.40 / 1.40 & 3.94 / 1.30 & 4.08 / 1.30 & 4.28 / 1.38 \\
& \quad HypLiLoc~\cite{Wang_2023_CVPR}   & CVPR'23 & 17 hrs.  & 5.03 / 1.46 & 4.31 / 1.43 & 3.61 / 1.11 & 2.61 / 1.09 & 3.89 / 1.27 \\
\multirow{-6}{*}{\rotatebox{90}{{APR}}} 
& \quad DiffLoc~\cite{Li_2024_CVPR} & CVPR'24 &  145 hrs.  & 2.03 / \underline{1.04} & 1.78 / \textbf{0.79} & 2.05 / \textbf{0.83} & 1.56 / \textbf{0.83} & 1.86 / \textbf{0.87} \\
& \quad FlashMix~\cite{Goswami_2025_WACV} & WACV'25 &  \textbf{1 hr.}  & 2.04 / 1.95 & 1.95 / 1.83 & 2.44 / 2.18 & 2.81 / 2.14 & 2.31 / 2.03 \\
\midrule
& \quad SGLoc~\cite{Li_2023_CVPR}   & CVPR'23 &  50 hrs.  & 1.79 / 1.67 & 1.81 / 1.76 & 1.33 / 1.59 & 1.19 / 1.39 & 1.53 / 1.60 \\
& \quad LiSA~\cite{Yang_2024_CVPR}  & CVPR'24 &  53 hrs.  & 0.94 / 1.10 & 1.17 / 1.21 & 0.84 / 1.15 & 0.85 / 1.11 & 0.95 / 1.14 \\
& \quad RALoc~\cite{Yang_2025_ICCV}  & ICCV'25 &  98 hrs.  & 1.53 / 1.48 & 1.72 / 1.27 & 1.38 / 1.21 & 1.42 / 1.24 & 1.51 / 1.30 \\
& \quad GTR-Loc~\cite{yugtr_2025_NIPS} & NeurIPS'25 & \underline{4 hrs.} & \textbf{0.77} / \textbf{1.02} & \underline{0.77} / 1.01 & \underline{0.67} / 1.01 & \underline{0.80} / 1.07 & \underline{0.75} / 1.03 \\
\multirow{-4}{*}{\rotatebox{90}{{SCR}}} 
& \quad LightLoc~\cite{Li_2025_CVPR}  & CVPR'25 &  \textbf{1 hr.}  & 0.82 / 1.12 & 0.85 / 1.07 & 0.81 / 1.11 & 0.82 / 1.16 & 0.83 / 1.12 \\
\cmidrule(lr){2-9}
& \quad LightLoc++  & Ours &  \textbf{1 hr.}  & \underline{0.79} / \underline{1.01} & \textbf{0.71} / \underline{0.94} & \textbf{0.62} / \underline{0.96} & \textbf{0.68} / \underline{0.99} & \textbf{0.70} / \underline{0.99} \\
\bottomrule
\end{tabular}
\label{tab:qeoxford}
	\vspace{0.2cm}  
\caption{\textbf{Quantitative results on the Oxford dataset.} Mean ATE (m)  /  ARE ($^{\circ}$) are reported.  Best results in \textbf{bold}, second best results \underline{underlined}. Training time is also included.}
\centering
\begin{tabular}{c|@{}l|c|c|cccc|c}
\toprule

& \quad Methods & Reference & Training Time  & {15-13-06-37} & {17-13-26-39} &  {17-14-03-00} & {18-14-14-42} & Average\\


%
\midrule
& \quad PointLoc~\cite{Wang_2022_Sensors}  & Sens. J.'22 &  125 hrs.  & 12.42 / 2.26 &  13.14 / 2.50 &  12.91 / 1.92 &  11.31 / 1.98 &  12.45 / 2.17 \\
& \quad PosePN~\cite{Yu_2022_PR}  & PR'22 &  5 hrs.  & 14.32 / 3.06 &  16.97 / 2.49 &  13.48 / 2.60 &  9.14 / 1.78 &  13.48 / 2.48  \\
& \quad PosePN++~\cite{Yu_2022_PR} & PR'22 &  11 hrs.  & 9.59 / 1.92 &  10.66 / 1.92 &  9.01 / 1.51 &  8.44 / 1.71 &  9.43 / 1.77 \\
& \quad STCLoc~\cite{Yu_2022_TITS}   & TITS'23 &  20 hrs.  & 6.93 / 1.48 &  7.55 / 1.23 &  7.44 / 1.24 &  6.13 / 1.15 &   7.01 / 1.28 \\
& \quad NIDALoc~\cite{Yu_2023_TITS} & TITS'24 &  20 hrs.  & 5.45 / 1.40 &  7.63 / 1.56 &  6.68 / 1.26 &  4.80 / 1.18 &   6.14 / 1.35\\
& \quad HypLiLoc~\cite{Wang_2023_CVPR}  & CVPR'23 &  17 hrs.  & 6.88 / 1.09 &  6.79 / 1.29 &  5.82 / \underline{0.97} &  3.45 / \underline{0.84} &   5.74 / 1.05\\
\multirow{-6}{*}{\rotatebox{90}{{APR}}} 
& \quad DiffLoc~\cite{Li_2024_CVPR} & CVPR'24 &  145 hrs.  & 3.57 / \textbf{0.88} &  3.65 / \textbf{0.68} &  4.03 / \textbf{0.70} &  2.86 / \textbf{0.60} &   3.53 / \textbf{0.72}\\
& \quad FlashMix~\cite{Goswami_2025_WACV} & WACV'25 &  \textbf{1 hr.}  & 3.05 / 1.96 & 4.55 / 2.05 & 4.67 / 2.05 & 2.94 / 1.79 & 3.80 / 1.96 \\
\midrule
& \quad SGLoc~\cite{Li_2023_CVPR}   & CVPR'23 &  50 hrs.  & 3.01 / 1.91 &  4.07 / 2.07 &  3.37 / 1.89 &  2.12 / 1.66 &  3.14 / 1.88 \\
& \quad LiSA~\cite{Yang_2024_CVPR} & CVPR'24 &  53 hrs.  & 2.36 / 1.29 & 3.47 / 1.43 & 3.19 / 1.34 & 1.95 / 1.23 & 2.74 / 1.32 \\
& \quad RALoc~\cite{Yang_2025_ICCV}  & ICCV'25 &  98 hrs.  & 3.19 / 4.10 & 3.87 / 3.96 & 3.32 / 3.87 & 2.59 / 3.71 & 3.24 / 3.91 \\
& \quad GTR-Loc~\cite{yugtr_2025_NIPS} & NeurIPS'25 & \underline{4 hrs.} & \underline{2.29} / 1.17 & \underline{3.07} / 1.21 & \textbf{2.99} / 1.20 & \underline{2.00} / 1.18 & \underline{2.59} / 1.19 \\
\multirow{-4}{*}{\rotatebox{90}{{SCR}}} 
& \quad LightLoc~\cite{Li_2025_CVPR}  & CVPR'25 &  \textbf{1 hr.}  & 2.33 / 1.21 & 3.19 / 1.34 & \underline{3.11} / 1.24 & 2.05 / 1.20 & 2.67 / 1.25 \\
\cmidrule(lr){2-9}
& \quad LightLoc++  & Ours &  \textbf{1 hr.}  & \textbf{2.24} / \underline{1.06} & \textbf{3.00} / \underline{1.04} & \textbf{2.99} / 1.04  & \textbf{1.90} / 1.01 & \textbf{2.53} / \underline{1.04} \\
\bottomrule
\end{tabular}
\label{tab:oxford}
\end{table*}

\subsection{Experimental Setup}
\noindent{\textbf{Evaluation Metrics.}}
We evaluate localization accuracy using two standard pose estimation metrics: Absolute Trajectory Error (ATE) and Absolute Rotation Error (ARE). ATE measures the Euclidean distance between the predicted and ground-truth translations, while ARE measures the angular difference between the predicted and ground-truth rotations:
\begin{equation}
\text{ATE} = \|\hat{\mathbf{t}} - \mathbf{t}\|_2, \quad\text{ARE} = \arccos\left(\frac{\text{trace}(\hat{\mathbf{R}}\mathbf{R}^\top)-1}{2}\right).
\end{equation}

For each trajectory with $N$ frames, we report the mean ATE and ARE over all frames. In addition, we report the training time to evaluate the efficiency of learning a new scene.

\noindent{\textbf{Implementation Details.}}
LightLoc++ is implemented using PyTorch~\cite{Paszke_2019_NIPS} and MinkowskiEngine~\cite{Choy_2019_CVPR}.

For sensor-robust backbone pretraining, we use the proposed SULID dataset. We apply K-Means clustering to the training positions and group spatially overlapping samples from different trajectories into 26 spatial scene partitions. Following LightLoc~\cite{Li_2025_CVPR}, each partition is assigned a dedicated regression head, while the backbone is shared across all partitions. 

Each training sample contains synchronized observations from multiple LiDAR sensors, which are transformed into a unified LiDAR coordinate system using calibrated LiDAR-to-LiDAR extrinsics. Each input point cloud is voxelized with a voxel size of 0.3 m before being fed into the backbone. 
For the $\mathcal{L}_{\text{global}}$, the distance threshold used to identify reliable cross-sensor correspondences is set to 0.3 m.
The backbone and all regression heads are jointly optimized using $\mathcal{L}_{\text{backbone}}$ in Eq.~\ref{eq_backbone}, with $\lambda=0.1$. We train for 100 epochs with a batch size of 128 using AdamW~\cite{Loshchilov_2020_ICLR}. More details can be found in the appendix.


We use mixed-precision training to reduce memory consumption, and backbone pretraining takes approximately 2 days on 4 NVIDIA GH200 GPUs.

For scene-specific heads training, the pretrained backbone is frozen, and only lightweight scene-specific prediction heads are optimized. For the Oxford and QEOxford dataset, sparse voxelization uses a voxel size of 0.3m, and the number of spatial clusters in SCG is set to $k=25$. A 150MB GPU feature buffer is maintained to cache global features for efficient SCG training. The SCG classification head is trained for 50 epochs with a batch size of 512. For RSD, the downsampling ratio $r_d$ is fixed at 0.25, while the start and stop ratios $r_{st}$ and $r_{sp}$ are set to 0.25 and 0.85, respectively. For SCR training, the regression head is optimized for 25 epochs with a batch size of 256 using AdamW~\cite{Loshchilov_2020_ICLR}. The learning rate follows a one-cycle schedule~\cite{Smith_2019_super}, ranging from $5\times10^{-4}$ to $5\times10^{-3}$. Dataset-specific settings, such as voxel size and feature buffer size, are provided in the appendix.

All scene-specific models are trained on a single NVIDIA RTX 4090 GPU, and all reported training times correspond to this setting for fair comparison.
\begin{table*}[h!t]
\caption{\textbf{Quantitative results on the Boreas dataset.} Mean ATE (m)  /  ARE ($^{\circ}$) are reported. . Best results in \textbf{bold}, second best results \underline{underlined}. Training time is also included.}
\centering
\begin{tabular}{c|@{}l|c|c|cccc|c}
\toprule

& \quad Methods & Reference & Training Time  & {2020-12-01} & {2021-02-02} & {2021-03-02} &  {2021-04-29}  & Average\\


%
\midrule
& \quad PointLoc~\cite{Wang_2022_Sensors}  & Sens. J.'22 &  26 hrs.  & 14.45 / 1.55 &  12.73 / 1.59  & 12.00 / 1.40 &  13.00 / 1.48  &  13.05 / 1.57 \\
& \quad PosePN~\cite{Yu_2022_PR}  & PR'22 &  6 hrs.  & 15.59 / 4.18 &  5.58 / 0.99 & 4.59 / 0.86 &  5.82 / 0.88  &  7.90 / 1.73  \\
& \quad PosePN++~\cite{Yu_2022_PR}   & PR'22 &  8 hrs.  & 7.69 / 1.45 &  4.87 / 0.98 & 4.53 / 0.89 &  5.14 / 0.87  &  5.56 / 1.01 \\
& \quad STCLoc~\cite{Yu_2022_TITS}   & TITS'23 &  13 hrs.  & 4.64 / 0.73 &  4.56 / 0.80 & 4.38 / 0.80 &  4.71 / 0.63  &   4.57 / 0.74 \\
& \quad NIDALoc~\cite{Yu_2023_TITS} & TITS'24 & 21 hrs.  & 5.01 / 0.87 &  4.87 / 0.82 & 4.87 / 0.82 &  4.78 / 0.71  &   4.88 / 0.81\\
& \quad HypLiLoc~\cite{Wang_2023_CVPR}   & CVPR'23 &  8 hrs.  & 4.46 / \underline{0.70} &  1.86 / 0.71  & 1.66 / \textbf{0.53} &  1.57 / \textbf{0.44}  &  2.39 / \textbf{0.60}\\
\multirow{-6}{*}{\rotatebox{90}{{APR}}} 
& \quad DiffLoc~\cite{Li_2024_CVPR} & CVPR'24 & 39 hrs.  & 6.70 / 3.05 &  5.64 / 2.86 &  5.40 / 2.41  &  4.58 / 2.15  &   5.58 / 2.62\\
& \quad FlashMix~\cite{Goswami_2025_WACV} & WACV'25 &  \textbf{1 hr.} & 52.67 / 12.63 & 2.36 / 1.90  & 2.35 / 1.78  & 1.91 / 1.49  & 14.82 / 4.45 \\
\midrule
& \quad SGLoc~\cite{Li_2023_CVPR}   & CVPR'23 &  15 hrs.  & 1.88 / 1.12 &  1.49 / 1.10 & 1.43 / 1.04 &  1.51 / 1.09  &  1.58 / 1.09 \\ 
& \quad LiSA~\cite{Yang_2024_CVPR}  & CVPR'24 &  18 hrs.  & 1.62 / 0.89 & \underline{1.34} / 0.86 & \underline{1.30} / 0.78  & \underline{1.41} / 0.81  & \underline{1.42} / 0.84 \\
& \quad RALoc~\cite{Yang_2025_ICCV}  & ICCV'25 &  43 hrs.  & \underline{1.58} / \underline{0.70} & 1.58 / \textbf{0.60} & 1.61 / \underline{0.58} & 1.59 / 0.63  & 1.59 / \underline{0.63} \\
& \quad GTR-Loc~\cite{yugtr_2025_NIPS} & NeurIPS'25 & \underline{4 hrs.}  & 2.51 / 1.21 & 2.36 / 1.14 & 2.31 / 1.06 & 2.37 / 1.14 & 2.39 / 1.14 \\
\multirow{-4}{*}{\rotatebox{90}{{SCR}}} 
& \quad LightLoc~\cite{Li_2025_CVPR}  & CVPR'25 &  \textbf{1 hr.}  & 2.39 / 1.18 & 2.24 / 1.14 & 2.20 / 1.04  & 2.33 / 1.12  & 2.29 / 1.12 \\
\cmidrule(lr){2-9}
& \quad LightLoc++  & Ours &  \textbf{1 hr.}  & \textbf{0.97} / \textbf{0.67} & \textbf{0.83} / \underline{0.65} & \textbf{0.78} / 0.59 & \textbf{0.88} / \underline{0.64} & \textbf{0.87} / 0.64 \\
\bottomrule
\end{tabular}
\label{tab:boreas}
\end{table*}

\subsection{Comparison with State-of-the-Art Methods}
\noindent{\textbf{Results on Oxford and QEOxford.}}
We first evaluate LightLoc++ on Oxford and QEOxford, both of which are collected using Velodyne HDL-32E LiDARs that are not included in the SULID pretraining set. We report the training time to assess scene adaptation efficiency, and use the mean ATE and mean ARE over all test trajectories to evaluate localization accuracy.

Tab.~\ref{tab:qeoxford} reports the results on QEOxford. LightLoc++ achieves 0.70m/0.99$^{\circ}$, improving upon LightLoc (0.83m/1.12$^{\circ}$) and achieving the best position accuracy and second-best orientation accuracy among all compared methods. Notably, LightLoc++ only replaces the backbone weights while preserving the original LightLoc architecture, enabling new-scene learning within one hour while maintaining the same compact 22M-parameter model. Compared with GTR-Loc, which introduces geospatial text supervision at the cost of approximately 4$\times$ longer training time, LightLoc++ achieves superior localization accuracy while retaining significantly higher training efficiency. It also outperforms the recently proposed RALoc. Although DiffLoc achieves slightly better orientation accuracy (0.87$^{\circ}$ vs. 0.99$^{\circ}$), LightLoc++ improves position accuracy by 62.4\% while requiring only single-frame input and 145$\times$ less training time. 

Tab.~\ref{tab:oxford} presents the evaluation results on the original Oxford benchmark. Consistent with the observations on QEOxford, LightLoc++ improves the average localization error of LightLoc from 2.67m/1.25$^{\circ}$ to 2.53m/1.04$^{\circ}$. It achieves the best position accuracy and the second-best orientation accuracy among all compared methods, while maintaining the one-hour training time. These results demonstrate that the proposed sensor-robust backbone improves localization performance on unseen LiDAR sensors without sacrificing the rapid deployment capability of LightLoc.

\noindent{\textbf{Results on Boreas.}}
We further evaluate the proposed method on the Boreas dataset, which is collected using a Velodyne Alpha-Prime 128-beam LiDAR. Compared with LightLoc, LightLoc++ achieves significant gains in both position and orientation accuracy, reaching 0.87m/0.64$^{\circ}$ and reducing the localization errors by 62.0\% and 42.9\%, respectively. Compared with LiSA, RALoc, GTR-Loc, and FlashMix, LightLoc++ achieves the best overall localization performance while retaining the fast deployment capability.

Since LightLoc++ keeps the downstream architecture and training protocol unchanged, the improvement over LightLoc on Boreas can be mainly attributed to the SULID-pretrained backbone. This demonstrates that multi-LiDAR consistency pretraining helps learn more sensor-robust representations, improving transferability to unseen LiDAR sensors.

\begin{table}[!t]
\caption{\textbf{Quantitative results on the MulRan dataset.} Mean ATE (m)  /  ARE ($^{\circ}$) are reported. . Best results in \textbf{bold}, second best results \underline{underlined}. Training time is also included.}
\centering
\begin{tabular}{c|@{}l|c|c|c}
\toprule

& \quad Methods & Reference & Training Time  & DCC\\


%
\midrule
& \quad PointLoc~\cite{Wang_2022_Sensors}  & Sens. J.'22 &  725 mins. & 16.38 / 5.16 \\
& \quad PosePN~\cite{Yu_2022_PR}  & PR'22 &  70 mins.  & 18.30 / 7.68 \\
& \quad PosePN++~\cite{Yu_2022_PR} & PR'22 &  200 mins.  & 6.64 / 3.43\\
& \quad STCLoc~\cite{Yu_2022_TITS}   &TITS'23 &  230 mins.  & 7.89 / 3.25\\
& \quad NIDALoc~\cite{Yu_2023_TITS} & TITS'24 &  321 mins.  & 5.87 / 3.39\\
& \quad HypLiLoc~\cite{Wang_2023_CVPR}   & CVPR'23 &  110 mins.  & 8.96 / 2.64\\
\multirow{-6}{*}{\rotatebox{90}{{APR}}} 
& \quad DiffLoc~\cite{Li_2024_CVPR} & CVPR'24 &  525 mins.  & 11.61 / 3.06\\
& \quad FlashMix~\cite{Goswami_2025_WACV} & WACV'23 &  \underline{20 mins.}  & 5.82 / 3.96\\
\midrule
& \quad SGLoc~\cite{Li_2023_CVPR}   & CVPR'23 &  203 mins  & 4.43 / 2.34\\
& \quad LiSA~\cite{Yang_2024_CVPR}  & CVPR'24 &  270 mins  & 4.59 / 3.17\\
& \quad RALoc~\cite{Yang_2025_ICCV}  & ICCV'25 &  577 mins  & \underline{3.42} / 2.64\\
& \quad GTR-Loc~\cite{yugtr_2025_NIPS} & NeurIPS'25 & 42 mins. & 3.51 / \textbf{2.13} \\
\multirow{-4}{*}{\rotatebox{90}{{SCR}}} 
& \quad LightLoc~\cite{Li_2025_CVPR}  & CVPR'25 &  \textbf{12 mins.}  & 4.76 / 2.77\\
\cmidrule(lr){2-5}
& \quad LightLoc++  & Ours &  \textbf{12 mins.}  & \textbf{3.36} / \underline{2.18}\\
\bottomrule
\end{tabular}
\label{tab:dcc}
\end{table}

\noindent{\textbf{Results on MulRan.}}
We further evaluate the proposed method on the DCC sequence of the MulRan dataset, which is collected by an Ouster OS1-64 LiDAR. As shown in Tab.~\ref{tab:dcc}, LightLoc++ achieves the best position accuracy among all compared methods, reaching 3.36m/2.18$^{\circ}$, and improves the position error of LightLoc from 4.76m to 3.36m while maintaining comparable orientation accuracy. Notably, LightLoc++ preserves the efficient deployment pipeline of LightLoc and requires only 12 minutes to learn a new scene, substantially faster than LiSA and RALoc.

\begin{table}[!t]
\caption{\textbf{Quantitative results on the HeRCULES dataset.} 
Mean ATE (m)  /  ARE ($^{\circ}$) are reported. 
The best and second-best results are highlighted in \textbf{bold} and \underline{underline}, respectively. Average training time across two scenes is also included.}
\centering
\begin{tabular}{c|@{}l|c|cc}
\toprule

& \quad Methods & Training Time & Library & Sports\\


%
\midrule
& \quad PointLoc~\cite{Wang_2022_Sensors}  &  288 mins.  & 6.25 / 3.15  &  4.45 / 4.62 \\
& \quad PosePN~\cite{Yu_2022_PR}  &  28 mins.  & 7.54 / 4.65  &  4.23 / 5.26   \\
& \quad PosePN++~\cite{Yu_2022_PR} &  70 mins.  & 4.70 / 2.75 &  5.82 / 4.93  \\
& \quad STCLoc~\cite{Yu_2022_TITS}   &  98 mins.  & 8.40 / 5.09  &  5.92 / 4.87  \\
& \quad NIDALoc~\cite{Yu_2023_TITS} &  113 mins.  & 3.15 / \underline{2.08}  &  2.45 / 3.66 \\
& \quad HypLiLoc~\cite{Wang_2023_CVPR}  &  39 mins.  & 5.81 / \textbf{2.07}  &  4.87 / 3.65 \\
\multirow{-6}{*}{\rotatebox{90}{{APR}}} 
& \quad DiffLoc~\cite{Li_2024_CVPR} &  211 mins.  & 13.69 / 6.15  &  12.27 / 14.21 \\
& \quad FlashMix~\cite{Goswami_2025_WACV} &  \textbf{11 mins.}  & 11.56 / 6.39  & 16.46 / 11.41 \\
\midrule
& \quad SGLoc~\cite{Li_2023_CVPR}   &  103 mins.  & 2.96 / 4.21  & 1.99 / 2.60\\
& \quad LiSA~\cite{Yang_2024_CVPR} &  132 mins.  & 2.48 / 2.40   & 1.73 / \textbf{3.04} \\
& \quad RALoc~\cite{Yang_2025_ICCV}  &  175 mins.  & \underline{2.13} / 2.60  & \textbf{1.69} / 3.49 \\
& \quad GTR-Loc~\cite{yugtr_2025_NIPS}  & 31 mins. & 3.19 / 3.23  & 2.94 / 3.92 \\
\multirow{-4}{*}{\rotatebox{90}{{SCR}}} 
& \quad LightLoc~\cite{Li_2025_CVPR} &  \underline{14 mins.}  & 2.66 / 2.58 & 1.88 / 3.26 \\
\cmidrule(lr){2-5}
& \quad LightLoc++  &  \underline{14 mins.}  & \textbf{1.79} / \underline{2.08}  & \underline{1.71} / \underline{3.11}\\
\bottomrule
\end{tabular}
\label{tab:hecules}
\end{table}

\noindent{\textbf{Results on HeRCULES.}}
Finally, we evaluate the proposed method on the HeRCULES dataset, which is collected using an Aeva Aeries II LiDAR. Compared with the rotating LiDAR sensors used in SULID pre-training, this sensor exhibits substantially different measurement characteristics, making HeRCULES a particularly challenging benchmark for evaluating representation generalization. As shown in Tab.~\ref{tab:hecules}, LightLoc++ achieves average localization errors of 1.75m/3.32$^{\circ}$ across the Library and Sports scenes, improving the position accuracy of LightLoc by 22.9\% while maintaining the same deployment efficiency. Compared with GTR-Loc, LightLoc++ reduces the average position error from 3.07m to 1.75m while requiring less than half of the training time. Although RALoc achieves a similar overall accuracy, it requires more than 12 times longer training time.

\begin{figure*}
  \centering
  \includegraphics[width=1\linewidth]{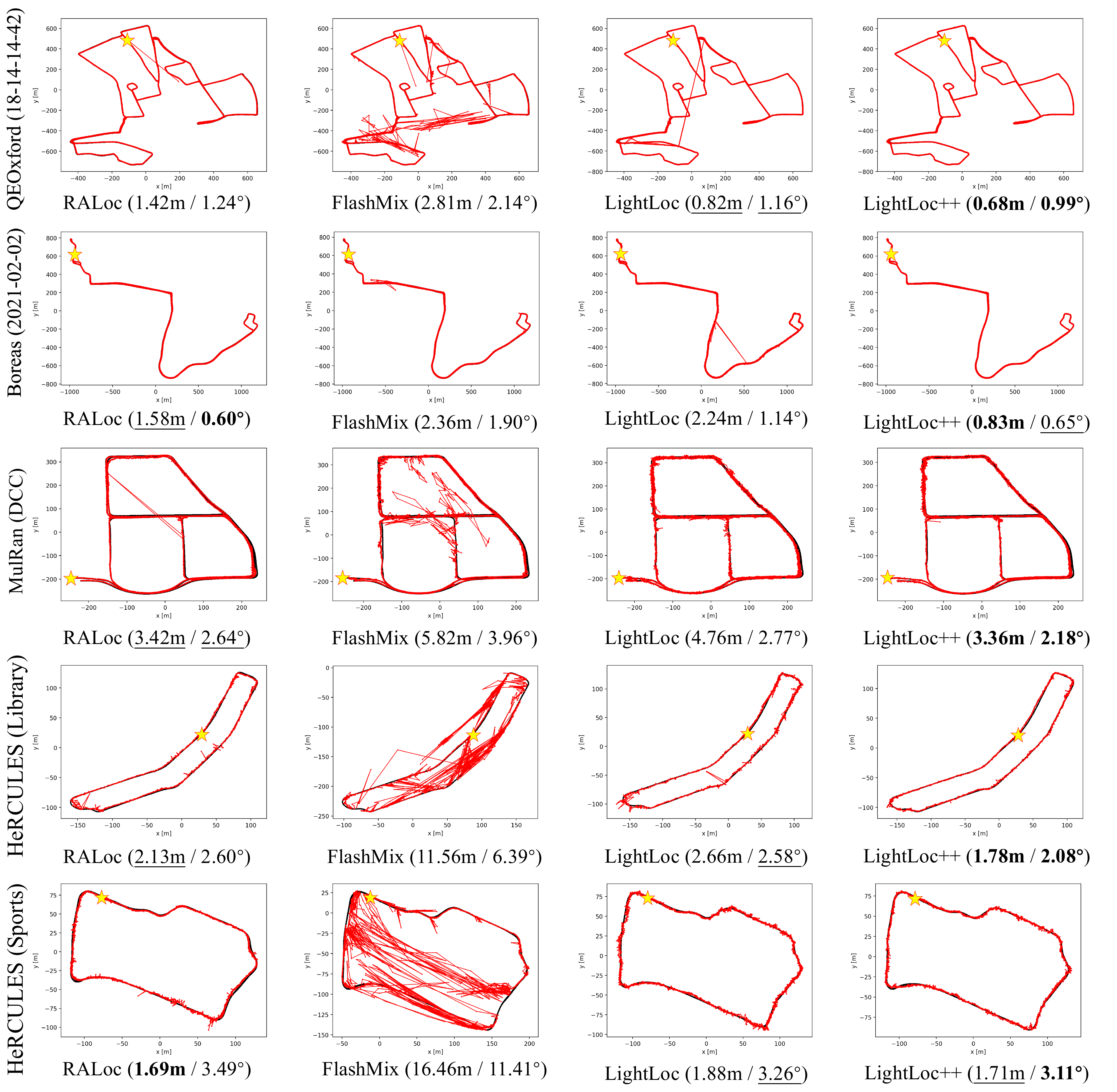}
\caption{\textbf{Visualization results of predicted trajectories on QEOxford, Boreas, MulRan, and HeRCULES.} 
The ground truth and prediction are black and red lines, respectively.
The star denotes the first frame. The caption of each subfigure shows the mean ATE (m) and ARE ($^{\circ}$). For each
trajectory, we highlight the best and second-best results.}
  \label{fig:vis}
\end{figure*}

These results demonstrate that our pretrained backbone can generalize to LiDAR sensors with substantially different sensing characteristics, enabling LightLoc++ to maintain strong localization accuracy and efficient new-scene learning under severe sensor domain shifts. As visualized in Fig.~\ref{fig:vis}, LightLoc++ generates smoother and more accurate trajectories than the compared methods, with predictions consistently closer to the ground-truth paths across different datasets.

\noindent{\textbf{Model Size and Runtime.}}
Model size, trainable parameters, and runtime are critical factors for practical LiDAR localization systems. LightLoc++ keeps the lightweight design of LightLoc, with 22M parameters in total, including a 16M-parameter backbone and 6M-parameter scene-specific heads. Since the backbone is frozen during new-scene learning, only the 6M parameters in the heads are optimized.

In contrast, LISA and SGLoc contain about 105M parameters, RALoc contains 104M, DiffLoc contains 40M, and HypLiLoc contains 52M, all of which require full model training for new scenes. FlashMix contains 56M parameters in total, with 15M trainable parameters for new-scene learning. GTR-Loc involves an 80M-parameter training framework, with 54M trainable parameters during new-scene learning. After distillation, it adopts the same inference structure as LightLoc. Therefore, LightLoc++ requires fewer trainable parameters while maintaining a lightweight model size and efficient inference. It runs at 25ms per scan, including regression and RANSAC-based pose refinement, corresponding to 50Hz and satisfying real-time deployment requirements.

\subsection{Analysis of Learned Representations}

\begin{table}[t]
\setlength{\tabcolsep}{5pt}
    \caption{\textbf{Effect of multi-LiDAR representation learning.} 
    We evaluate localization performance across multiple LiDAR benchmarks using encoders pretrained under different settings. Results are reported as mean ATE (m) / ARE ($^\circ$).}
	\resizebox{\linewidth}{!}{
	\begin{tabular}{@{}l|cccc}
		\toprule
		  Pretraining & QEOxford  & Boreas  & MulRan & HeRCULES            \\ \hline
		  KITTI &1.50 / 1.73  & 2.75 / 1.17 & 5.33 / 4.32  & 2.09 / 2.98 \\  
        nuScenes & 0.83 / 1.12  & 2.12 / 1.24    & 4.76 / 2.77 & 2.27 / 2.92  \\ 
        HeliPR  &0.95 / 1.27  & 1.31 / 0.82    & 3.62 / 2.50 & 2.13 / 2.95\\  \hline  
        SULID &0.72 / 1.03    &0.98 / \underline{0.66}    &\underline{3.50} / \underline{2.39} & 2.02 / 2.81 \\
        +$\mathcal{L}_{\text{global}}$ & \textbf{0.69} / \underline{1.02}   & \underline{0.92} / \underline{0.66}    & 3.56 / 2.41 & \underline{1.83} / \underline{2.62} \\
        +$\mathcal{L}_{\text{local}}$ & \underline{0.70} / \textbf{0.99}   & \textbf{0.87} / \textbf{0.64}    & \textbf{3.36} / \textbf{2.18} & \textbf{1.75} / \textbf{2.60} \\
		\bottomrule
	\end{tabular}
	}
    \label{tab:encoder}
\end{table}

\noindent{\textbf{Effect of Multi-LiDAR Pretraining.}}
Tab.~\ref{tab:encoder} compares different pretraining datasets under the same downstream localization framework, where the backbone is frozen and only the scene-specific regression heads are optimized. This setting allows us to directly evaluate the generalization ability of the learned LiDAR representations.

We first compare two single-sensor pretraining datasets, KITTI and nuScenes. Overall, nuScenes provides stronger performance than KITTI, especially on QEOxford, Boreas, and MulRan. This suggests that larger-scale urban driving data helps improve the generalization ability of the pretrained backbone. However, both datasets are collected using a single LiDAR type, which limits their robustness.

We then compare HeLiPR, which contains heterogeneous LiDAR sensors and uses Ouster and Aeva LiDAR data in this paper. HeLiPR improves over nuScenes on Boreas and MulRan, reducing the ATE from 2.12m to 1.31m and from 4.76m to 3.62m, respectively. This indicates that multi-LiDAR data is beneficial for learning more generalizable representations. However, HeLiPR does not consistently improve over nuScenes on all benchmarks, showing that sensor diversity alone is not sufficient.

Using SULID for pretraining further improves performance across all benchmarks. Compared with HeLiPR, SULID reduces the localization error from 0.95m/1.27$^{\circ}$ to 0.72m/1.03$^{\circ}$ on QEOxford, from 1.31m/0.82$^{\circ}$ to 0.98m/0.66$^{\circ}$ on Boreas, from 3.62m/2.50$^{\circ}$ to 3.50m/2.39$^{\circ}$ on MulRan, and from 2.13m/2.95$^{\circ}$ to 2.02m/2.81$^{\circ}$ on HeRCULES. In this setting, SULID is used only as a multi-LiDAR pretraining dataset, without explicit cross-sensor consistency constraints. These results suggest that, beyond sensor diversity, large-scale data collection and richer coverage of typical urban scenarios are important for learning generalizable representations.

\noindent{\textbf{Effect of Cross-Sensor Consistency Constraints.}}
Starting from the SULID-pretrained backbone, we further exploit the synchronized multi-LiDAR observations by introducing cross-sensor consistency constraints. 

As shown in Tab.~\ref{tab:encoder}, adding the global consistency loss $\mathcal{L}_{\text{global}}$ improves the performance on most benchmarks. In particular, it reduces the ATE from 0.72m to 0.69m on QEOxford, from 0.98m to 0.92m on Boreas, and from 2.02m to 1.83m on HeRCULES. This shows that aligning scene-level representations across synchronized LiDAR sensors helps the backbone learn a more sensor-robust representation.

Further adding the local consistency loss $\mathcal{L}_{\text{local}}$ leads to the best overall performance. Compared with the SULID-pretrained backbone without consistency constraints, the final model reduces the localization error from 0.72m/1.03$^{\circ}$ to 0.70m/0.99$^{\circ}$ on QEOxford, from 0.98m/0.66$^{\circ}$ to 0.87m/0.64$^{\circ}$ on Boreas, from 3.50m/2.39$^{\circ}$ to 3.36m/2.18$^{\circ}$ on MulRan, and from 2.02m/2.81$^{\circ}$ to 1.75m/2.60$^{\circ}$ on HeRCULES. This demonstrates that local point-level consistency complements global scene-level alignment for exploiting synchronized multi-LiDAR observations.

\subsection{Cross-Sensor Generalization}
\begin{table}[t]
\centering
\caption{\textbf{Cross-Sensor Generalization.}
LiSA, RALoc, and FlashMix are included as reference localization methods to indicate the cross-sensor generalization capability of existing approaches.
For our framework, SCG and RSD are disabled, and only the pretrained encoder is varied.
Results are reported as mean ATE (m) / ARE ($^\circ$).}
\label{tab:encoder_generalization}
\setlength{\tabcolsep}{2pt}
\resizebox{\linewidth}{!}{
\begin{tabular}{l|ccc|c}
\toprule
Method & O$\rightarrow$O & O$\rightarrow$A & O$\rightarrow$V & Average \\
\midrule

\multicolumn{5}{c}{\textbf{Reference Localization Methods}} \\
\midrule

LiSA      & 0.88 / 0.70 & \underline{2.18} / \underline{2.62} & 101.94 / 47.56 & 41.39 / 14.35 \\
RALoc     & 1.95 / 1.01 & 22.05 / 26.01 & 246.13 / 120.23 & 90.04 / 49.08 \\
FlashMix  & 6.76 / 5.69 & 245.61 / 73.77 & 164.89 / 100.68 & 104.32 / 60.05 \\

\midrule
\multicolumn{5}{c}{\textbf{Our Framework with Different Pretrained Backbones}} \\
\midrule

KITTI                & 1.74 / 1.14 & 38.22 / 25.63 & 213.73 / 103.25 & 84.56 / 43.34  \\
nuScenes             & 0.97 / 0.64 & 40.31 / 20.58 & 186.27 / 85.23 & 75.85 / 35.48  \\
HeliPR               & \textbf{0.69} / \underline{0.50} & \textbf{1.40} / \textbf{1.51} & 34.20 / 20.14 & 12.10 / 7.38  \\ \hline
SULID                & 0.72 / \underline{0.50} & 2.46 / \underline{2.75} & 3.18 / \underline{3.53} & \underline{2.12} / \underline{2.26}  \\ 
+$\mathcal{L}_{\text{global}}$ & 0.72 / \textbf{0.49} & 2.41 / 2.83 & \underline{3.15} / 3.61 & 2.09 / 2.31  \\
+$\mathcal{L}_{\text{local}}$  & \underline{0.71} / \underline{0.50} & 2.29 / \underline{2.75} & \textbf{1.94} / \textbf{3.06} & \textbf{1.65} / \textbf{2.10}  \\

\bottomrule
\end{tabular}}
\label{tab:generalization}
\end{table}
\begin{figure}[t]
  \centering
  \includegraphics[width=1\columnwidth]{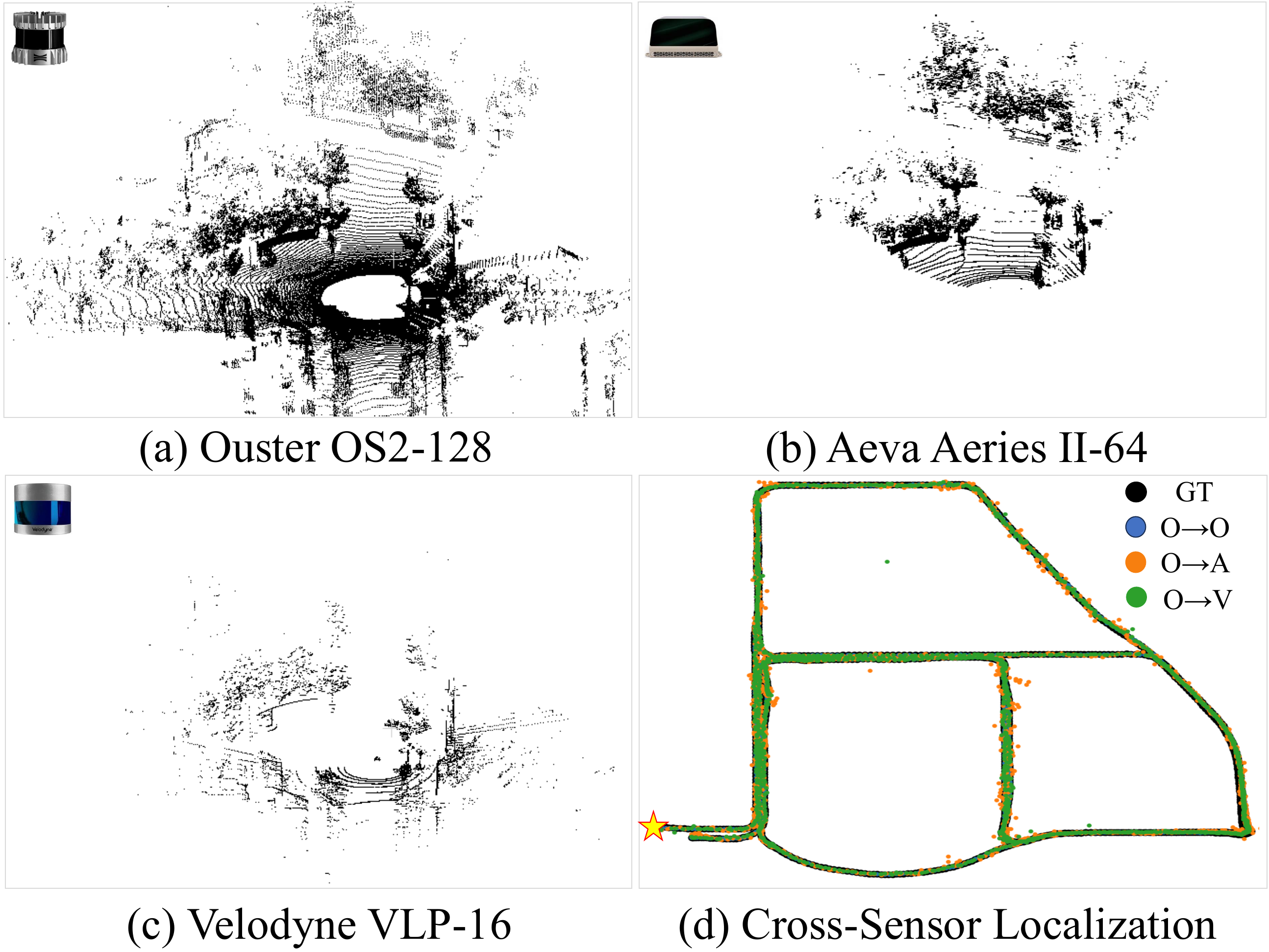}
  \caption{
\textbf{Cross-sensor localization generalization on the HeLiPR DCC sequence.}
Despite the substantial differences in sensing characteristics and fields of view, our method maintains accurate localization across Ouster, Aeva, and Velodyne sensors without fine-tuning. The
star denotes the first frame.  
}
   \label{fig:cross}
\end{figure}
We further evaluate cross-sensor generalization on the DCC sequence of HeLiPR. In this setting, all methods are trained only with Ouster OS2-128 data and are directly tested on Ouster OS2-128 (O), Aeva Aeries II (A), and Velodyne VLP-16 (V) data without any fine-tuning. This protocol evaluates whether a localization model trained on one LiDAR sensor can generalize to other sensors observing the same scene geometry.

For LightLoc++, we freeze the pretrained backbone and train only the regression head with Ouster OS2-128 data. To isolate the effect of representation learning, SCG and RSD are disabled, and only the pretrained backbone is varied. This allows us to directly examine whether the learned backbone can produce consistent and generalizable representations.

Fig.~\ref{fig:cross}~(a)-(c) visualize point clouds collected by O, A, and V at the same location. Although the sensors observe the same scene, their point clouds show clear differences in sampling patterns, fields of view, and point densities. In particular, V produces much sparser observations than O and A, making the O$\rightarrow$V setting more challenging for cross-sensor localization.

The first three rows of Tab.~\ref{tab:generalization} report the results of reference localization methods. Under the O$\rightarrow$O setting, all methods achieve reasonable performance, indicating that they can perform well when the training and testing data come from the same LiDAR sensor. In the O$\rightarrow$A setting, FlashMix suffers from a significant performance drop, while LiSA still maintains reasonable accuracy. This suggests that the semantic cues used by LiSA retain certain robustness under moderate sensor changes. However, under the more challenging O$\rightarrow$V setting, all reference methods fail to localize reliably.

We then evaluate our framework with different pretrained backbones. When using KITTI and nuScenes pretraining, showing a similar trend as reference methods, performance drops significantly in the O$\rightarrow$A and O$\rightarrow$V setting. 
HeLiPR pretraining achieves strong results on O$\rightarrow$O and O$\rightarrow$A, partly because Ouster and Aeva sensors are included during its backbone pretraining. 
However, its performance drops significantly under O$\rightarrow$V, where Velodyne is not included in pretraining. 
In contrast, the SULID-pretrained backbone achieves much lower error on O$\rightarrow$V, reducing the ATE/ARE from 34.20m/20.14$^{\circ}$ to 3.18m/$^{\circ}$ compared with HeLiPR. This shows that SULID pretraining provides more generalizable LiDAR representations, especially for unseen sensor types.

Finally, we evaluate the effect of the cross-sensor consistency learning. Compared with the SULID-pretrained backbone, adding the consistency constraints further improves cross-sensor generalization, especially under the most challenging O$\rightarrow$V setting. The ATE/ARE is reduced from 3.18m/3.53$^{\circ}$ to 1.94m/3.06$^{\circ}$, and the average error is reduced from 2.12m/2.26$^{\circ}$ to 1.65m/2.10$^{\circ}$. This demonstrates that exploiting synchronized multi-LiDAR observations during pretraining helps the backbone learn more sensor-robust LiDAR representations. Such representations produce more consistent responses to the same scene geometry under different LiDAR sampling patterns, reducing the impact of sensor discrepancies.

\subsection{Effect of SCG and RSD on Training Efficiency}

\begin{table}[t]
	\caption{\textbf{Ablation study}. Training time required to achieve comparable accuracy on QEOxford. \textbf{Backbone}: multi-LiDAR backbone. \textbf{SCG}: sample classification guidance for regression learning. \textbf{RSD}: redundant sample downsampling for filtering well-learned samples during training.}
		\centering
		\resizebox{1\linewidth}{!}
  {
		\begin{tabular}{c|ccc|cc}
			\toprule
			             & Backbone            &  SCG         & RSD     & Training Time     & Mean Error~[m/$^{\circ}$]\\
			\midrule
			1            &               &               &               & 5.7 hrs.                 & 1.02 / 1.23\\
			2            & $\checkmark$  &               &               & 2.4 hrs.                 & 0.84 / 1.12\\
			3            & $\checkmark$    & $\checkmark$	  & 	 	      & 1.3 hrs.            & 0.71 / 0.95\\
			4            & $\checkmark$    &               & $\checkmark$    & 2 hrs.             & 0.86 / 1.15\\
            5            & $\checkmark$    & $\checkmark$    & $\checkmark$    & 1 hrs.            & 0.70 / 0.99\\
			\bottomrule
		\end{tabular}
	}
    \label{tab:ablation}
\end{table}

Tab.~\ref{tab:ablation} evaluates the impact of the sensor-robust backbone, SCG, and RSD on new-scene learning. Starting from random initialization, training the regression model requires 5.7 hours and achieves a localization error of 1.02 m/1.23$^{\circ}$. Replacing it with the proposed pretrained backbone reduces the training time to 2.4 hours and improves the accuracy to 0.84m/1.12$^{\circ}$. This shows that the learned backbone provides strong fixed representations for efficient scene-specific head training.

\noindent{\textbf{Effect of SCG.}} 
Building upon the pretrained backbone, we further evaluate the effect of SCG. As shown in Rows~2 and~3 of Tab.~\ref{tab:ablation}, SCG improves both ATE and ARE, reducing the localization error from 0.84m/1.12$^{\circ}$ to 0.71m/0.95$^{\circ}$. Meanwhile, the training time is reduced from 2.4 hours to 1.3 hours. This demonstrates that the sample-wise probability distribution learned by SCG provides effective spatial guidance for regression, improving both localization accuracy and training efficiency. Notably, SCG requires only 5 minutes of auxiliary training for each new scene.

We further study the sensitivity of the cluster number $k$ in Tab.~\ref{tab:scg}. The classification accuracy remains above 99\% across a wide range of cluster numbers, indicating that the learned scene partition is stable. In practice, we set $k=25$, which achieves the best localization accuracy and is used in all experiments.

\begin{table}[t]
    \caption{\textbf{Effect of cluster number $k$ in SCG.}We report the classification accuracy (\%) and localization error (ATE/ARE) on the QEOxford dataset.}
	\resizebox{\linewidth}{!}{
	\begin{tabular}{@{}l|ccccc}
		\toprule
		Clusters & 5 & 15 & 25  & 50          & 100            \\ \hline
		Acc. &100 &99.89 & 99.76 & 99.62  & 99.17 \\ 
        Err. & 0.76/1.06  & 0.73/1.05  & 0.70/0.99    & 0.73/1.02 & 0.77/1.07      \\
		\bottomrule
	\end{tabular}
	}
    \label{tab:scg}
\end{table}
\begin{table}[t]
    \centering
    \caption{\textbf{Impact of different sampling strategies and ratios}. We report the mean error [m/$^{\circ}$] on the QEOxford dataset.}
	\begin{tabular}{@{}l|ccc}
		\toprule
		Sampling & Random          & Uniform        & RSD           \\ \hline
		15$\%$  & 0.84 / 1.21 & 0.77 / 0.98   & 0.71 / 0.97     \\
        25$\%$  & 0.96 / 1.27 & 0.83 / 1.00   & 0.70 / 0.99     \\
        35$\%$  & 1.12 / 1.43 & 1.04 / 1.32   & 0.82 / 1.02     \\
		\bottomrule
	\end{tabular}
    \label{tab:rsd}
\end{table}

\noindent{\textbf{Effect of RSD.}} 
The results for RSD are shown in Row 4 and Row 5 of Tab.~\ref{tab:ablation}. Comparing Row 2 and Row 4, RSD reduces training time from 2.4 hours to 2 hours with minimal accuracy loss (0.84m/1.12$^{\circ}$ vs. 0.86m/1.15$^{\circ}$). Additionally, comparing Row 3 and Row 5, when using our method with SCG, the improvement is significant, boosting training efficiency by 23.1\%. With RSD, training time is further reduced, ultimately achieving the goal of completing training within 1 hour. This study demonstrates that RSD can significantly reduce training time without sacrificing accuracy, effectively addressing the challenge of handling large volumes of data in large-scale outdoor scenes.

We also compare RSD with random and uniform sampling under different pruning ratios, as shown in Tab.~\ref{tab:rsd}. Uniform sampling achieves comparable performance at a 15\% pruning ratio, but its accuracy degrades when more samples are removed. In contrast, RSD introduces no measurable accuracy drop at 25\% pruning and remains effective even at 35\%. By identifying well-learned samples through prediction variance, RSD reduces redundant training samples while preserving informative supervision.
\section{Conclusion}
In this paper, we presented LightLoc++, a sensor-robust and efficient outdoor LiDAR localization framework. We first show that the efficiency of decoupled SCR methods largely depends on the generalization ability of the pretrained backbone, which can degrade under different LiDAR configurations. To address this issue, we introduced SULID, a synchronized urban multi-LiDAR dataset covering representative LiDAR types and diverse urban scenes. Based on SULID, we adopt global and local cross-sensor consistency learning to improve the sensor robustness of the backbone. LightLoc++ further preserves efficient new-scene learning by freezing the backbone and optimizing only lightweight scene-specific heads, with SCG and RSD reducing regression ambiguity and redundant computation. Extensive experiments on benchmark datasets demonstrate that LightLoc++ achieves strong localization accuracy, efficient training, and robust cross-sensor generalization across different LiDAR sensors.

\section*{Appendix}
\appendices

\subsection{Benchmark Datasets}
\label{sup_datasets}

\begin{table}[t]
	\centering
	\centering
    \caption{Dataset Descriptions on the Oxford and QEOxford dataset.}
		\begin{tabular}{c|cc|cc}
			\toprule
			Sequence     &Length	& Tag      &Training     &Testing  \\ \hline
			11-14-02-26  &9.37km	& Sunny    & $\checkmark$  &      \\
			14-12-05-52  &9.22km	& Overcast & $\checkmark$  &      \\
			14-14-48-55  &9.04km	& Overcast & $\checkmark$  &      \\
			18-15-20-12  &9.04km	& Overcast & $\checkmark$  &	     \\
			15-13-06-37  &8.85km	& Overcast & 			 &$\checkmark$   \\
			17-13-26-39  &9.02km	& Sunny    &  			 &$\checkmark$   \\
			17-14-03-00  &9.02km	& Sunny    &  			 &$\checkmark$   \\
			18-14-14-42  &9.04km	& Overcast &   		  	 &$\checkmark$   \\
			\bottomrule
		\end{tabular}
	\label{sup:oxford}
	\vspace{0.1cm}  
	\centering
    \caption{Dataset Descriptions on the Boreas dataset.}
	\resizebox{1\linewidth}{!}{
    \setlength{\tabcolsep}{2pt}
		\begin{tabular}{c|cc|cc}
			\toprule
			Sequence     &Length	& Tag      &Training     &Testing  \\ \hline
			2020-11-26   &7.95km	& Overcast, Snow, Loop    & $\checkmark$  &      \\
			2021-03-23  &7.95km	& Overcast, Construction, Loop & $\checkmark$  &      \\
			2021-03-30  &7.95km	& Sunny, Cloudy, Construction, Loop & $\checkmark$  &      \\
			2021-04-13  &7.96km	& Sunny, Cloudy, Construction, Loop & $\checkmark$  &	     \\
			2020-12-01  &7.95km	& Overcast, Snow, Snowing, Loop & 			 &$\checkmark$   \\
			2021-02-02  &7.94km	& Overcast, Snow, Loop    &  			 &$\checkmark$   \\
			2021-03-02  &7.97km	& Sunny, Cloudy, Snow, Loop    &  			 &$\checkmark$   \\
			2021-04-29  &7.95km	& Overcast, Rain, Loop &   		  	 &$\checkmark$   \\
			\bottomrule
		\end{tabular}
	}
	\label{sup:boreas}
    \vspace{0.1cm}  
	\centering
    \caption{Dataset Descriptions on the DCC Sequence of MulRan dataset.}
		\begin{tabular}{c|cc|cc}
			\toprule
			Sequence     &Length	& Tag      &Training     &Testing  \\ \hline
			DCC01  &4.92km	& Sunny    & $\checkmark$  &      \\
			DCC02  &5.43km	& Sunny & $\checkmark$  &      \\ 
			DCC03  &5.44km	& Sunny & 			 &$\checkmark$   \\
			\bottomrule
		\end{tabular}
	\label{sup:mulran}
    \vspace{0.1cm}  
	\centering
    \caption{Dataset Descriptions on the Sports and Library Sequences of HeRCULES Dataset.}
		\begin{tabular}{c|cc|cc}
			\toprule
			Sequence     &Length	& Tag      &Training     &Testing  \\ \hline
			  Sports01   &1.40km	& Sunny   & $\checkmark$  &      \\
			Sports02  &0.71km	& Cloudy & $\checkmark$  &      \\
			Sports03  &1.40km	& Snow &   &$\checkmark$	     \\    \hline
			Library01  &1.51km	& Sunny    &$\checkmark$ 			 &   \\
			Library02  &1.52km	& Cloudy    &$\checkmark$  			 &   \\
			Library03  &0.76km	& Snow &   		  	 &$\checkmark$   \\
			\bottomrule
		\end{tabular}
	\label{sup:hercules}
    \vspace{0.1cm}  
	\centering
    \caption{Dataset Descriptions on the DCC Sequence of HeLiPR dataset.}
		\begin{tabular}{c|cc|cc}
			\toprule
			Sequence     &Length	& Tag      &Training     &Testing  \\ \hline
			DCC04  &5.51km	& Sunny, Night    & $\checkmark$  &      \\
			DCC06  &4.65km	& Sunny, Night  & $\checkmark$  &      \\ 
			DCC05  &5.31km	& Sunny, Morning & 			 &$\checkmark$   \\
			\bottomrule
		\end{tabular}
	\label{sup:helipr}
\end{table}

We use Oxford Radar RobotCar~\cite{Dan_2020_ICRA}, QEOxford~\cite{Li_2023_CVPR}, Boreas~\cite{burnett2023boreas}, MulRan~\cite{kim2020mulran}, and HeRCULES~\cite{kim2025hercules} for standard localization benchmarking. HeLiPR~\cite{jung_2024_helipr} is additionally used for cross-sensor generalization evaluation. Detailed sequence splits and conditions are provided below.

\noindent{\textbf{Oxford Radar RobotCar (Oxford).}}
The Oxford Radar RobotCar~\cite{Dan_2020_ICRA} dataset was collected in January 2019 along a central Oxford route using dual Velodyne HDL-32E LiDARs. It contains repeated traversals under different weather and illumination conditions, such as sunny and overcast scenes, making localization challenging. Following previous works~\cite{Li_2023_CVPR,Li_2024_CVPR,Li_2025_CVPR,yugtr_2025_NIPS,Yang_2025_ICCV}, we use the left LiDAR for evaluation. Specifically, we use 11-14-02-26, 14-12-05-52, 14-14-48-55, and 18-15-20-12 for training, and 15-13-06-37, 17-13-26-39, 17-14-03-00, and 18-14-14-42 for testing.

\noindent{\textbf{QEOxford.}} The QEOxford~\cite{Li_2023_CVPR} dataset is a quality-enhanced version of the Oxford dataset, where the GPS/INS pose errors are reduced through trajectory alignment. The improved pose quality provides more reliable supervision for scene coordinate regression-based localization. Following SGLoc~\cite{Li_2023_CVPR}, QEOxford uses the same training and testing split as Oxford.

\noindent{\textbf{Boreas}.}
The Boreas~\cite{burnett2023boreas} dataset was mainly collected along a repeated route near the University of Toronto with a Velodyne AlphaPrime 128-beam LiDAR, referred to as the Glen Shields route. The data were collected over the course of one year, covering diverse seasonal and weather conditions, such as sun, rain, and snow. In our experiments, we use four sequences, 2020-11-26, 2021-03-23, 2021-03-30, and 2021-04-13, for training, and four sequences, 2020-12-01, 2021-02-02, 2021-03-02, and 2021-04-29, for testing. The sequence lengths and weather tags are summarized in Tab.~\ref{sup:boreas}.

\noindent{\textbf{MulRan}.}
The MulRan~\cite{kim2020mulran} DCC sequence was collected around the Daejeon Convention Center in South Korea using an Ouster OS1-64 LiDAR. Although the DCC route is relatively short, it contains diverse urban structures, including a square, narrow roads between high-rise buildings, a mountain area, and crossroads.
Following FlashMix~\cite{Goswami_2025_WACV}, we use DCC01 and DCC02 for training and DCC03 for testing. The trajectory lengths and weather tags are summarized in Tab.~\ref{sup:mulran}.

\noindent{\textbf{HeRCULES}.}
The HeRCULES~\cite{kim2025hercules} dataset contains heterogeneous LiDAR measurements collected with an Aeva Aeries II sensor. In this work, we use the Library and Sports Complex sequences. The Library sequence was collected along a long and narrow one-way campus road near the library, including curves as well as uphill and downhill sections. The Sports Complex sequence was collected around a sports complex, covering parking areas and roads with flat, gently sloped, and steep sections. For the Sports scene, Sports01 and Sports02 are used for training, and Sports03 is used for testing. For the Library scene, Library01 and Library02 are used for training, and Library03 is used for testing. The trajectory lengths and weather tags are summarized in Tab.~\ref{sup:hercules}.

\noindent{\textbf{HeLiPR}.}
HeLiPR~\cite{jung_2024_helipr} is a heterogeneous LiDAR dataset designed for inter-LiDAR place recognition under spatiotemporal variations. In this work, we use the DCC trajectories for cross-sensor generalization evaluation. Specifically, DCC04 and DCC06 are used for training, while DCC05 is used for testing. The regression head is trained only with Ouster OS2-128 data and directly evaluated on Ouster OS2-128, Aeva Aeries II, and Velodyne VLP-16 data without fine-tuning. This setting allows us to evaluate both same-sensor localization performance and cross-sensor robustness. The trajectory lengths and conditions are summarized in Tab.~\ref{sup:helipr}.
\subsection{Sensor-Robust Backbone Training}
\label{sup_backbone}

\begin{figure}[t]
  \centering
  \includegraphics[width=1\columnwidth]{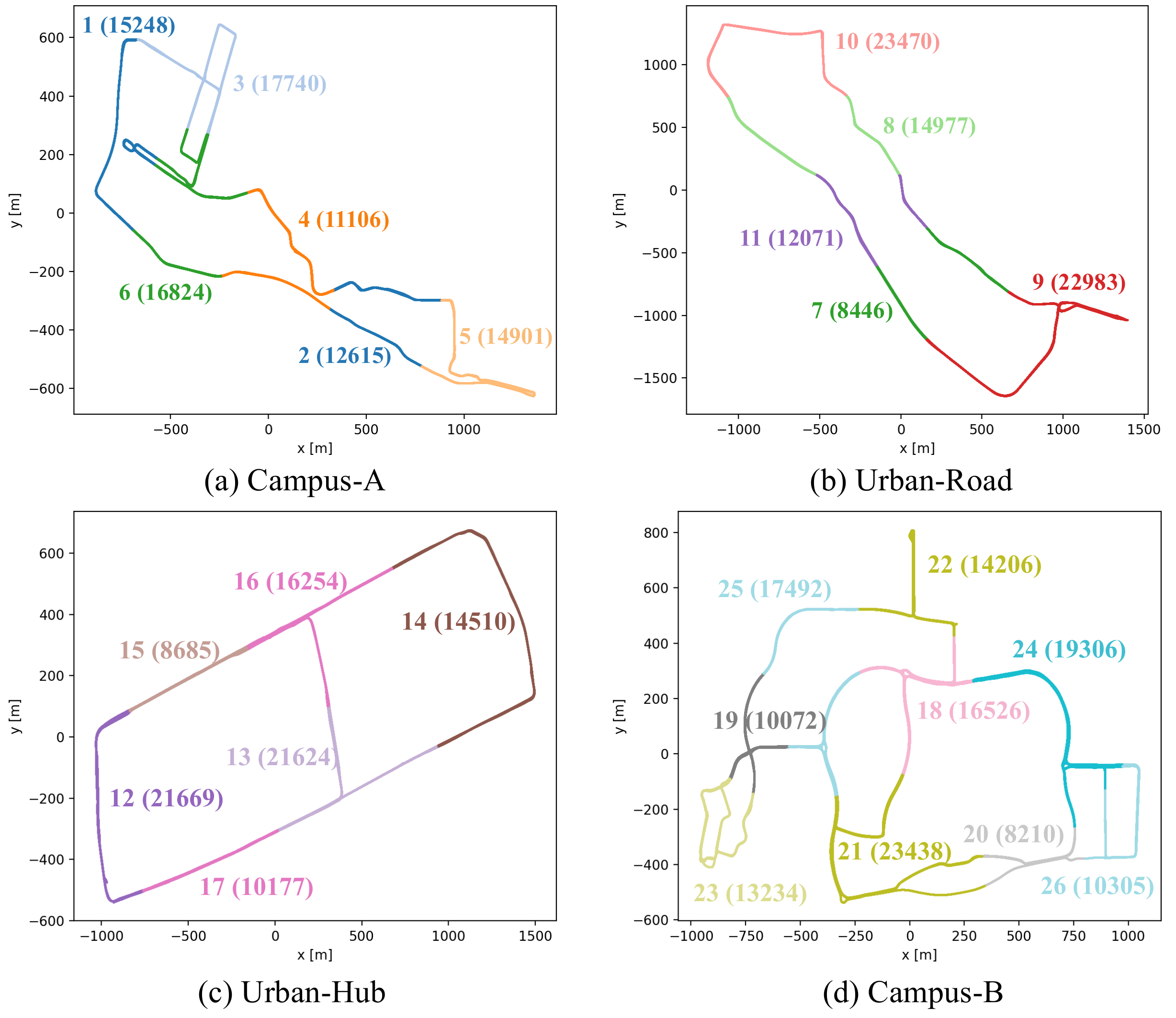}
   \caption{\textbf{Visualization of the K-Means-based spatial partitions in SULID.} Different colors denote different scene partitions, and the number in parentheses indicates the number of synchronized groups.}
   \label{fig:sup_split}
\end{figure}

We apply K-Means clustering to the training positions and divide SULID into 26 spatial scene partitions. As shown in Fig.~\ref{fig:sup_split}, each partition contains a set of spatially overlapping samples from different trajectories, and the number of samples in each partition is reported in the figure. Following LightLoc~\cite{Li_2025_CVPR}, each partition is assigned a dedicated regression head, while the backbone is shared across all partitions.

Each training sample consists of synchronized observations from multiple LiDAR sensors, which are transformed into the Ouster OS1-128 LiDAR coordinate system using calibrated LiDAR-to-LiDAR extrinsics. Each input point cloud is voxelized with a voxel size of 0.3 m before being fed into the backbone. 
For the local consistency loss, the distance threshold for selecting reliable cross-sensor correspondences is set to 0.3 m.
Each regression head is implemented as a 6-layer MLP with a hidden dimension of 512 and a skip connection after the third layer. During pretraining, the backbone and all regression heads are jointly optimized using $\mathcal{L}_{\text{backbone}}$, where the regression loss provides scene-coordinate supervision and the cross-sensor consistency loss encourages the backbone to produce similar representations for the same scene geometry under different LiDAR sampling patterns.

We use mixed-precision training and apply random data augmentation to the input point clouds. With a probability of 0.5, we apply translations along the $x$ and $y$ axes within $[-1,1]$ m, roll and pitch rotations within $[-5^\circ,5^\circ]$, and yaw rotations within $[-10^\circ,10^\circ]$. The scene-coordinate labels remain unchanged under these input perturbations. We train the backbone for 100 epochs with a batch size of 128 using AdamW~\cite{Loshchilov_2020_ICLR}. The learning rate follows a single-cycle cosine annealing schedule with a 5-epoch linear warm-up and a peak learning rate of $1\times10^{-3}$.
\subsection{Scene-Specific Prediction Heads}
\label{sup_heads}
After sensor-robust backbone pretraining, the backbone is frozen and only scene-specific prediction heads are optimized for each target dataset. For standard localization benchmarks, we use dataset-specific voxel sizes: 0.3m for Oxford, QEOxford, and HeRCULES, 0.4m for MulRan, and 0.5m for Boreas. We also set the GPU feature buffer size according to the dataset scale, with 160K samples for Oxford and QEOxford, 50K for Boreas, 20K for MulRan, and 10K for HeRCULES. The number of spatial clusters in SCG is fixed to $k=25$ for all datasets.

For classification head training, we accelerate the process by maintaining a GPU buffer that stores global features and their corresponding classification labels. The buffer is filled by repeatedly cycling through the shuffled training sequence, where each point cloud is augmented using the same data augmentation strategy as backbone pretraining. The classification head is then trained by iterating over the shuffled buffer. In practice, SCG training takes about 5 minutes for each new scene, including the time for filling the buffer, and completes 50 epochs with a batch size of 512.

For regression head training, we also apply the same data augmentation strategy as backbone pretraining to each input point cloud. The augmented point cloud is then passed through the frozen backbone to obtain dense descriptors. The features from the classification head are normalized to the unit sphere, perturbed with Gaussian noise, and normalized again. The dense descriptors and classification features are concatenated, and 256 voxels are randomly selected to regress their corresponding scene coordinates. The regression head is trained for 25 epochs on Oxford, QEOxford, Boreas, and HeRCULES, and for 15 epochs on MulRan. We use a batch size of 256 for all datasets except Boreas, where the batch size is set to 100. For RSD, we use the same setting across all datasets: the downsampling ratio $r_d$ is set to 0.25, and the start and stop ratios $r_{st}$ and $r_{sp}$ are set to 0.25 and 0.85, respectively.

{\small
\bibliographystyle{IEEEtran}
\bibliography{main}
}

\end{document}